\documentclass{article}

\usepackage{authblk}
\usepackage{amsmath}
\usepackage{graphicx}
\usepackage{hyperref}
\newcommand*{\email}[1]{\href{mailto:#1}{\nolinkurl{#1}} } 

\newcommand*{\alg}{\textsf{ENAD}} 

\usepackage{physics}
\usepackage{mathtools}
\usepackage{xcolor}
\usepackage[normalem]{ulem}
\usepackage[font=footnotesize]{caption}
\usepackage{subcaption}
\usepackage{algorithm}
\usepackage{algorithmicx}
\usepackage{algpseudocode}
\usepackage{amsthm}
\usepackage[backend=bibtex]{biblatex}
\DeclareMathOperator*{\argmax}{arg\,max}
\DeclareMathOperator*{\argmin}{arg\,min}
\DeclareMathOperator{\sign}{sign}
\DeclareMathOperator{\clip}{clip}

\algrenewcommand\algorithmicrequire{\textbf{Input:}}
\algrenewcommand\algorithmicensure{\textbf{Output:}}
\algnewcommand\algorithmicforeach{\textbf{for each}}
\algdef{S}[FOR]{ForEach}[1]{\algorithmicforeach\ #1\ \algorithmicdo}

\newcommand{\softmax}{\ensuremath{\sigma_{\text{sm}}}}
\newcommand{\logit}{\ensuremath{\sigma_{\text{lgt}}}}

\newcommand{\labelphantom}[1]{%
  \parbox{0pt}{\phantomsubcaption\label{#1}}%
}

\begin{document}

\title{Unity is strength: improving the detection of adversarial examples with ensemble approaches}

\author[1]{Francesco Craighero}
\author[1]{Fabrizio Angaroni}
\author[1]{Fabio Stella}
\author[2,3]{Chiara Damiani}
\author[1,3,4]{Marco Antoniotti}
\author[5,1,3]{Alex Graudenzi}

\affil[1]{Dept. of Informatics, Systems and Communication, Universit\`{a} degli Studi di Milano-Bicocca, Milan, Italy}

\affil[2]{Dept. of Biotechnology and Biosciences, Universit\`{a} degli Studi di Milano-Bicocca, Milan, Italy}

\affil[3]{B4 - Bicocca Bioinformatics Biostatistics and Bioimaging Centre, Universit\`{a} degli Studi di Milano-Bicocca, Milan, Italy}

\affil[4]{Bioinformatics Program, Tandon School of Engineering, NYU Poly, New York, NY, U.S.A}

\affil[5]{Institute of Molecular Bioimaging and Physiology, National Research Council (IBFM-CNR), Milan, Italy}

\affil[ ]{\email{f.craighero@campus.uinimib.it}, \email{fabrizio.angaroni@unimib.it}, \email{fabio.stella@unimib.it}, \email{chiara.damiani@unimib.it}.
\email{marco.antoniotti@unimib.it}, \email{alex.graudenzi@unimib.it}}

\date{}

\maketitle

\abstract{A key challenge in computer vision and deep learning is the definition of robust strategies for the detection of adversarial examples. 
Here, we propose the adoption of ensemble approaches to leverage the effectiveness of multiple detectors in exploiting distinct properties of the input data.
To this end, the ENsemble Adversarial Detector (\alg{}) framework integrates scoring functions from state-of-the-art detectors based on Mahalanobis distance, Local Intrinsic Dimensionality, and One-Class Support Vector Machines, which process the hidden features of deep neural networks. \alg{} is designed to ensure high standardization and reproducibility to the computational workflow.

Importantly, extensive tests on benchmark datasets, models and adversarial attacks show that \alg{} outperforms all competing methods in the large majority of settings. 
The improvement over the state-of-the-art and the intrinsic generality of the framework, which allows one to easily extend \alg{} to include any set of detectors, set the foundations for the new area of ensemble adversarial detection.}

\section*{Introduction}

\begin{figure}
\begin{center}
\includegraphics[width=\textwidth]{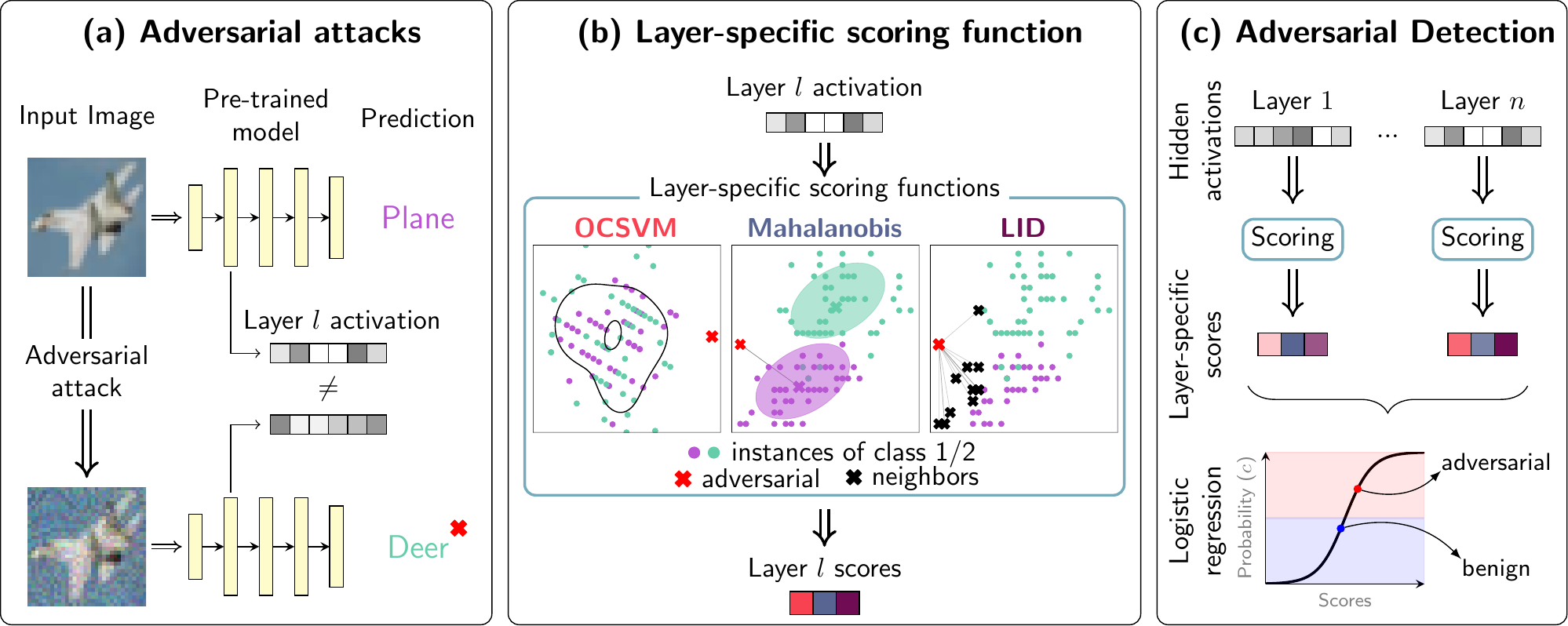}
\end{center}
   \caption{\footnotesize{{\bfseries \alg{} framework for adversarial detection}. A schematic depiction of the \alg{} framework is displayed. \textbf{(a)} Given an input image, which can be either benign or adversarial, and a pre-trained deep neural network, the activations of the hidden layers are extracted. \textbf{(b)} In order to measure the distance of the image with respect to training examples, layer-specific scores are computed via functions based either on One-class Support Vector Machines, Mahalanobis distance \cite{leeSimpleUnifiedFramework2018} or Local Intrinsic Dimensionality  \cite{maCharacterizingAdversarialSubspaces2018}. \textbf{(c)} Layer- and detector-specific scores are integrated via logistic regression, so to classify the image as benign or adversarial, with a confidence $c$.}}
\label{fig:framework}
\end{figure}

Deep Neural Networks (DNNs) have achieved impressive results in complex machine learning tasks, in a variety of fields such as computer vision \cite{krizhevskyImageNetClassificationDeep2017} and computational biology \cite{seniorImprovedProteinStructure2020}.

However, recent studies have shown that state-of-the-art DNNs for object recognition tasks are vulnerable to \emph{adversarial examples} \cite{szegedyIntriguingPropertiesNeural2014, goodfellowExplainingHarnessingAdversarial2015}. For instance, in the field of computer vision, adversarial examples are perturbed images that are misclassified by a given DNN, even if being almost indistinguishable from the original (and correctly classified) image. 
Adversarial examples have been investigated in many additional real-world applications and settings, including malware detection \cite{kolosnjajiAdversarialMalwareBinaries2018} and speech recognition \cite{qinImperceptibleRobustTargeted2019}.

Thus, understanding and countering adversarial examples has become a crucial challenge for the widespread adoption of DNNs in safety-critical settings, and resulted in the development of an ever-growing number of \emph{defensive techniques}. 
Among the possible countermeasures, some aims at increasing the robustness of the DNN model during the training phase, via \emph{adversarial training} \cite{szegedyIntriguingPropertiesNeural2014, goodfellowExplainingHarnessingAdversarial2015} or \emph{defensive distillation} \cite{papernotDistillationDefenseAdversarial2016} (sometimes referred to as \emph{proactive} methods). 
Alternative approaches aim at \emph{detecting} adversarial examples in the test phase, by defining specific functions for their detection and filtering-out (\emph{reactive} methods).

In this paper, we introduce a novel \emph{ensemble approach} for the detection of adversarial examples, named \textsf{EN}semble \textsf{A}dversarial \textsf{D}etector (\alg{}), which integrates scoring functions computed from multiple detectors that process the hidden layer activation of pre-trained DNNs. 
The underlying rationale is that, given the high-dimensionality of the hidden layers of Convolutional Neural Networks (CNNs) and the hardness of the adversarial detection problem, different algorithmic strategies might be effective in capturing and exploiting distinct properties of adversarial examples. Accordingly, their combination might allow one to better tackle the classical trade-off between generalization and overfitting, outperforming single detectors, as already suggested in \cite{aggarwalOutlierEnsemblesIntroduction2017} in the distinct context of outlier identification.
To the best of our knowledge, this is the first time that an ensemble approach is applied to the hidden features of DNNs for adversarial detection, while recently a combination of multiple detectors have been proposed for out-of-distribution detection \cite{kaurDetectingOODsDatapoints2021}. 

In detail, \alg{} includes two state-of-the-art detectors, based on Mahalanobis distance \cite{leeSimpleUnifiedFramework2018} and Local Intrinsic Dimensionality (LID) \cite{maCharacterizingAdversarialSubspaces2018}, and a newly developed detector based on One-Class SVMs (OCSVMs) \cite{scholkopfSupportVectorMethod1999}. 
OCSVMs were previously adopted for adversarial detection in \cite{maNICDetectingAdversarial2019}, but we extended the previous method by defining both a pre-processing step and a Bayesian hyperparameter optimization strategy, both of which result fundamental to achieve performances comparable to \cite{maCharacterizingAdversarialSubspaces2018, leeSimpleUnifiedFramework2018}. In the current implementation, the output of each detector is then integrated via a simple logistic regression that returns both the adversarial classification and the overall confidence of the prediction. A schematic depiction of the \alg{} framework is provided in Figure \ref{fig:framework}. 

Importantly, \alg{} is designed to ensure a high reproducibility standard of the computational workflow. 
In fact, similarly to \cite{raghuramGeneralFrameworkDetecting2021}, we defined the adversarial detection problem by: (i) clarifying the design choices related to data partitioning and feature extraction, and (ii) explicitly defining the detector-specific scoring functions and the way in which they are integrated. As a consequence, the overall framework of \alg{} is completely general and might be easily extended to any arbitrary set of detectors.

For the sake of reproducibility, the performance of \alg{} and competing methods was assessed with the extensive setting originally proposed in \cite{leeSimpleUnifiedFramework2018} and, in particular, we performed experiments with two models, namely ResNet \cite{heDeepResidualLearning2016} and  DenseNet \cite{huangDenselyConnectedConvolutional2017}, trained on CIFAR-10 \cite{krizhevskyLearningMultipleLayers}, CIFAR-100 \cite{krizhevskyLearningMultipleLayers} and SVHN \cite{netzerReadingDigitsNatural2011}, and considering four benchmark adversarial attacks, i.e., FGSM \cite{goodfellowExplainingHarnessingAdversarial2015}, BIM \cite{kurakinAdversarialExamplesPhysical2017}, DeepFool \cite{moosavi-dezfooliDeepFoolSimpleAccurate2016} and CW \cite{carliniAdversarialExamplesAre2017}.

\section*{Results}
\label{sec:results}

We compared the performance of \alg{} with stand-alone detectors A (OCSVM), B (Mahalanobis \cite{leeSimpleUnifiedFramework2018}), and C (LID \cite{maCharacterizingAdversarialSubspaces2018}). 
All four detectors integrate layer-specific anomaly scores via logistic regression and classify any example as adversarial if the posterior probability is $>0.5$, benign otherwise. More details are available in the Methods.

All approaches were tested on $2$ networks architectures (ResNet and DenseNet), $3$ benchmark datasets (CIFAR-10, CIFAR-100 and SVHN) and $4$ adversarial attacks (FGSM, BIM, DeepFool and CW), for a total of $24$ distinct configurations, as originally proposed in \cite{leeSimpleUnifiedFramework2018}.
For the evaluation of the performances, we employed two standard threshold-independent metrics, namely the Area Under the Receiver Operating Characteristic (AUROC) and the Area Under Precision Recall (AUPR) \cite{davisRelationshipPrecisionrecallROC2006}, which are detailed in the Methods section.

\begin{figure}[t]
\begin{center}
\includegraphics[width=\linewidth]{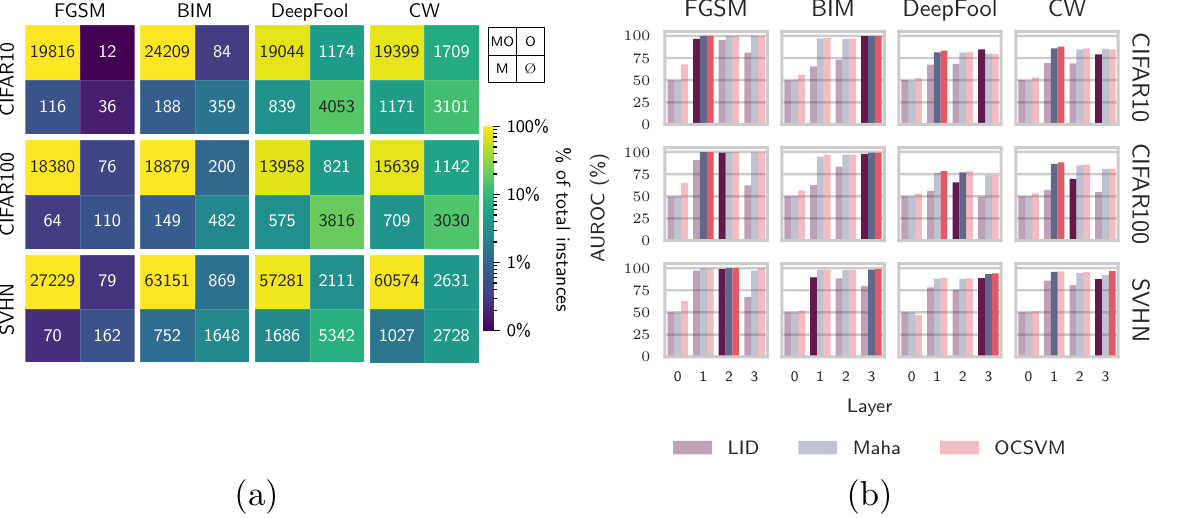}
\end{center}
\labelphantom{fig:heatmap}
\labelphantom{fig:layer-imp}
\caption{\footnotesize{{\bfseries (a) Comparison of predictions of single detectors (OCSVM vs. Mahalanobis -- DenseNet).} The contingency
table shows the number of adversarial examples of the test set $\mathcal L^{test}$ correctly identified by: both the OCSVM and the Mahalanobis detectors (OM), either one of the two methods (O or M), none of them ($\emptyset$). The results of the DenseNet model, with respect to the distinct datasets and attacks are shown, whereas the remaining pairwise comparisons are displayed in the Supplementary Material. {\bfseries (b) Influence of the hidden layers in adversarial detection (DenseNet).} For each configuration of datasets and attacks on the DenseNet model, the AUROC of each layer-specific score for detectors A--C is returned.  For each configuration and detector, the best performing layer is highlighted with a darker shade (see Methods for further details).}}
\end{figure}

\subsection*{Stand-alone Detectors Capture Distinct Properties of Adversarial Examples}

In order to assess the ability of detectors A, B and C to exploit different properties of input instances, we first analyzed the methods as stand-alone, and computed the subsets of adversarial examples identified: $(i)$ by all detectors, $(ii)$ by a subset of them, $(iii)$ by none of them.

In Figure \ref{fig:heatmap}, we reported a contingency table in which we compare the OCSVM and Mahalanobis detectors on all the experimental settings with the DenseNet model, while the remaining pairwise comparisons are presented in the Supplementary Material. 
Importantly, while the class of examples identified by both approaches is, as expected, the most crowded, we observe a substantial number of instances that are identified by either one of the two approaches. This important result appear to be general, as it is confirmed in the other comparisons between stand-alone detectors. 

In addition, in Supplementary Figure 4 one can find the layer-specific scores returned by all detectors in a specific setting (ResNet, DeepFool, CIFAR-10). For a significant portion of examples, the ranking ordering among scores is not consistent across detectors, confirming the distinct effectiveness in capturing different data properties in the hidden layers. 

To investigate the importance of the layers with respect to the distinct attacks, models and datasets, we also computed the AUROC directly on the layer-specific scores, i.e. the anomaly scores returned by each detector in each layer.
In Figure \ref{fig:layer-imp}, one can find the results for all detectors in all settings, with the DenseNet model. 
For the FGSM attack, the scores computed on the middle layers consistently return the best AUROC in all datasets, while for the BIM attack the last layer is apparently the most important. Notably, with DeepFool and CW attacks the most important layers are dataset-specific.
This result demonstrates that each attack may be vulnerable in distinct layers of the network.  

\subsection*{\alg{} Outperforms Stand-alone Detectors}

Table~\ref{tab:results} reports the AUROC and AUPR computed on the fitted adversarial posterior probability for detectors A, B, and C, respectively, on all $24$ experimental settings.

It can be noticed that \alg{} exhibits both the best AUROC and the best AUPR in $22$ out of $24$ settings (including $1$ tie with OCSVM), with the greatest improvements emerging in the hardest attacks, i.e. DeepFool and CW. 
Remarkably, the newly designed OCSVM detector outperforms the other stand-alone detectors in 12 and 14 settings in terms of AUROC and AUPR, respectively.

Notice that in Supplementary Table~5, we also evaluated the performance of all pairwise combinations of the three detectors, so to quantitatively investigate the impact of integrating the different algorithmic approaches, proving that distinct ensembles of detectors can be effective in specific experimental settings. 

\begin{table}
    \centering
    \includegraphics[width=0.8\textwidth]{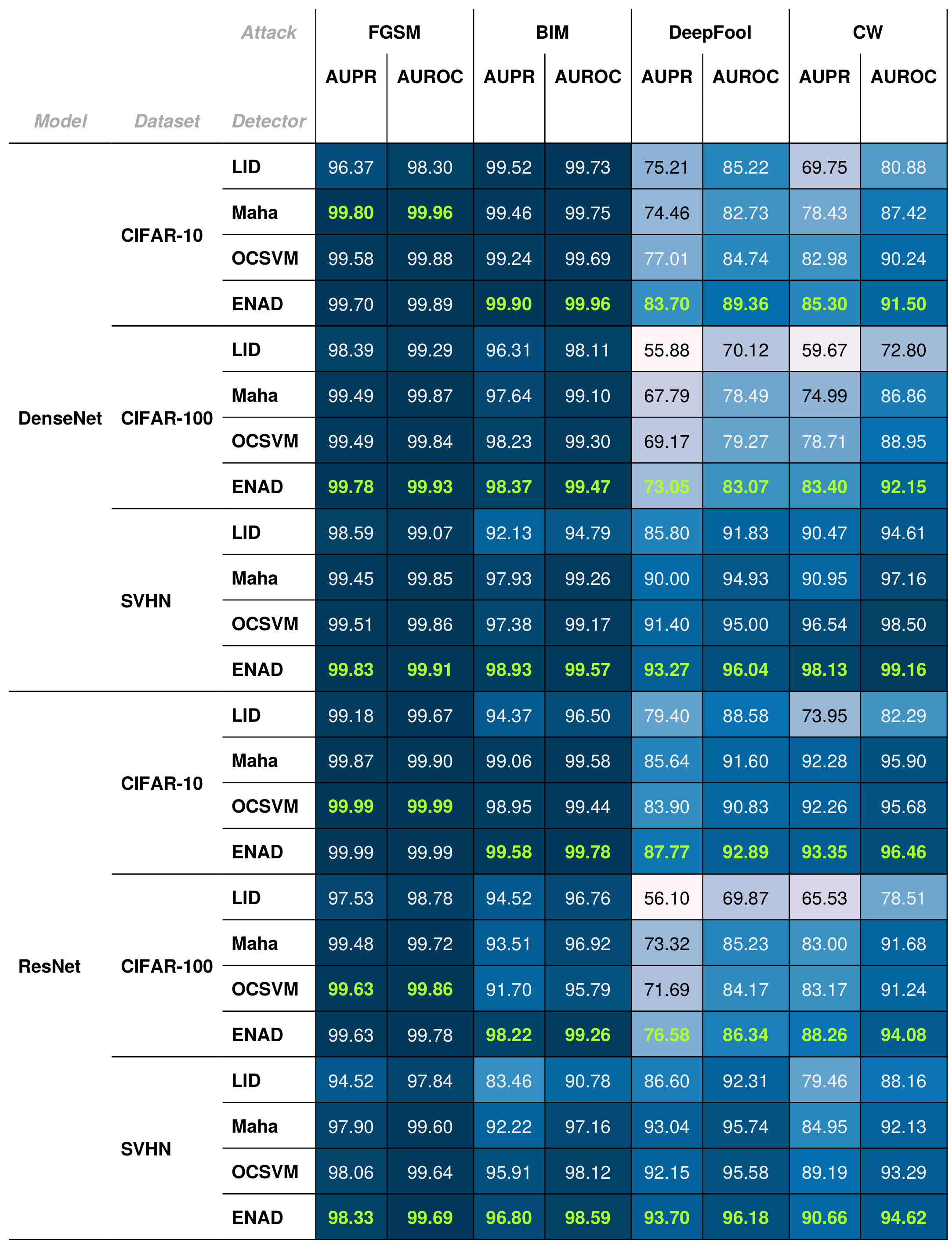}
    \caption{\footnotesize{\textbf{Comparative assessment of \alg{} and competing methods.} Performance comparison of the \alg{},  LID \cite{maCharacterizingAdversarialSubspaces2018}, Mahalanobis  \cite{leeSimpleUnifiedFramework2018}, OCSVM detectors (all the pairwise combinations of the three single detectors are available in Supplementary Table~5). The Table contains the AUROC and AUPR for all the combinations of selected datasets (CIFAR-10, CIFAR-100 and SVHN), models (DenseNet and ResNet), and attacks (FGSM, BIM). See Methods for further details.}}
    \label{tab:results}
\end{table}

\begin{figure}
\begin{center}
\includegraphics[width=\textwidth]{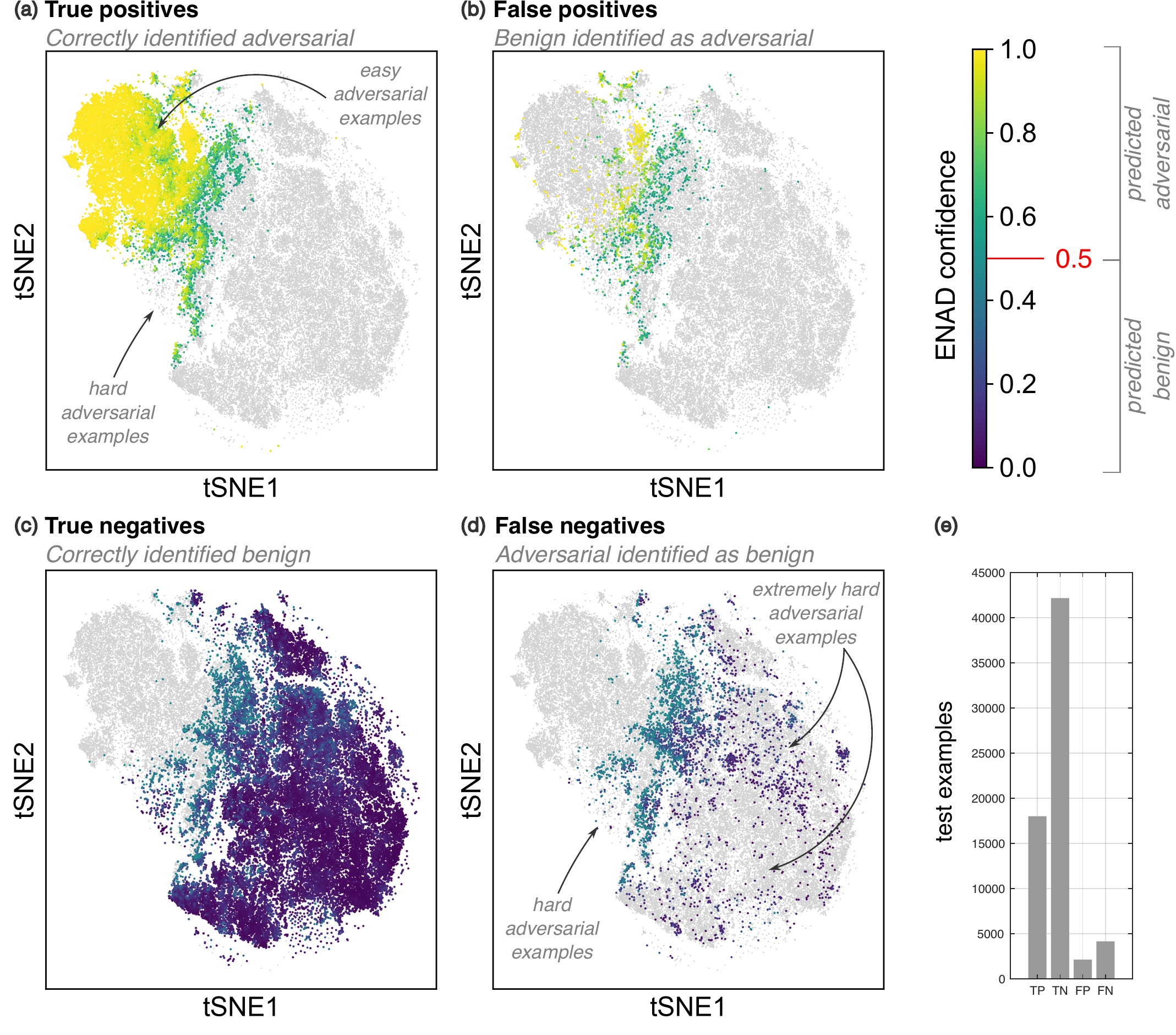}
\end{center}
   \caption{\footnotesize{{\bfseries Visualization of adversarial and benign examples in the low-dimensional score space (DenseNet -- SVHN -- DeepFool).} Representation of the test data $\mathcal L^{test}$ for the configuration DeepFool, DenseNet and SVHN in the tSNE low-dimensional space \cite{vandermaatenVisualizingDataUsing2008}. The layer- and detector-specific scores are weighted with the logit weights, Z-scored, and then used as features for the tSNE computation, via the computation of the k-nearest neighbour graph ($k = 50$) with Pearson correlation as metric (further tSNE parameters: perplexity = 100, early exaggeration = 100, learning rate = 10000). In each quadrant the true positives \textbf{(a)},  false positives \textbf{(b)},  true negatives \textbf{(c)}  and  false negatives \textbf{(d)} are displayed. Each point in the plot represents either an adversarial or a benign example and the color returns the confidence $c$ provided by \alg{}, i.e., the probability returned by the logistic classifier. \textbf{(e)} The bar-plots return the absolute number of TPs, TNs, FPs and FNs for this experimental setting.}}
\label{fig:tsne}
\end{figure}

\subsection*{Visualizing test examples in the low-dimensional score space}
In order to explore the relation between the scoring functions and the overall performance of \alg{}, it is possible to visualize the test examples on the low-dimensional tSNE space \cite{vandermaatenVisualizingDataUsing2008}, using the (logit weighted and Z-scored) layer- and detector-specific scores as starting features ($3$ detectors $\times$ $4$ hidden layers $ = 12$ initial dimensions). 

This representation allows one to intuitively assess how similar the score profiles of the test examples are: closer data points are those displaying more similar score profiles, which translates in an analogous distance from the set of correctly classified training instances, with respect to the three scoring functions currently included in \alg{}. 
Importantly, this allows one to evaluate how many and which adversarial examples display score profiles closer that those of benign ones, and vice versa, and visualize them. 

As an example, in Figure \ref{fig:tsne} the test set of the SVHN, DenseNet, DeepFool setting is displayed on the tSNE space.  
The color gradient returns the confidence $c$ of \alg{}, i.e., the probability of the logistic regression: any example is categorized as adversarial if $c > 0.5$, benign otherwise. 

While most of the adversarial examples are identified with high confidence (leftmost region of the tSNE plot), a narrow region exists in which adversarial examples overlap with benign ones, hampering their identification and leading to significant rates of both false positives and false negatives. 
Focusing on the set of false negatives, it is evident that some adversarial examples are scattered in the midst of the set of benign instances (rightmost region of the tSNE plot), rendering their identification extremely difficult. 

\subsection*{\alg{} and OCSVM detectors are robust against unknown attacks}
\label{sec:unknown}
We finally assessed the performance of \alg{} and detectors A, B, and C when an unknown attack is performed. 
In detail, we performed the hyperparameters optimization and the logit fit on the FGSM attack, for all methods, and then tested the performance on BIM, DeepFool and CW attacks. 
Despite the expected worsening of the performances, either \alg{} or OCSVM achieve the best AUROC and AUPR in almost all settings (see Table \ref{tab:unknown}), proving their robustness and applicability to the real-world scenarios in which the adversarial example type might be unknown. 

\begin{table}
    \centering
    \includegraphics[scale=.48]{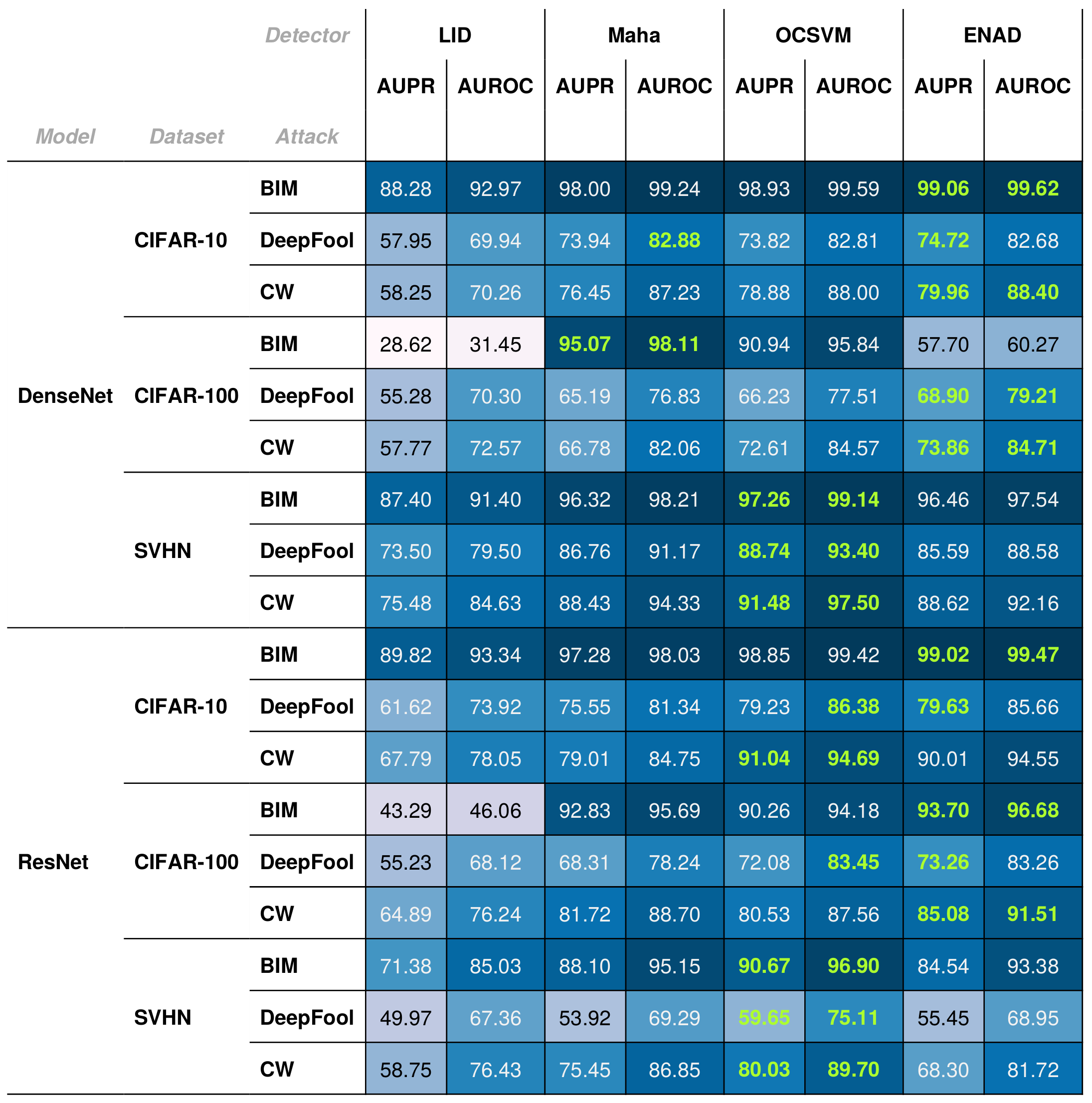}
    \caption{\footnotesize{\textbf{Comparative assessment in the unknown attack scenario.}} Performance comparison of the \alg{},  LID \cite{maCharacterizingAdversarialSubspaces2018}, Mahalanobis  \cite{leeSimpleUnifiedFramework2018}, and OCSVM detectors when both the hyperparameter optimization and the logistic regression fit are performed on the FGSM attack. The Table contains the AUROC for all the combinations of selected datasets (CIFAR-10, CIFAR-100 and SVHN), models (DenseNet and ResNet), and attacks (BIM, DeepFool and CW). See Methods for further details. }
    \label{tab:unknown}
\end{table}

\section*{Conclusions}
We introduced the \alg{} ensemble approach for adversarial detection, motivated by the observation that distinct detectors are able to isolate non-overlapping subsets of adversarial examples, by exploiting different properties of the input data in the internal representation of a DNN.
Accordingly, the integration of layer-specific scores extracted from three independent detectors (LID, Mahalanobis and OCSVM) allows \alg{} to achieve significantly improved performance on benchmark datasets, models and attacks, with respect to the state-of-the-art, even when the simplest integration scheme (i.e., logit) is adopted. 

It is also worth of note that the newly introduced OCSVM detector proved highly effective as stand-alone in our tests, indicating that the use of  one-class classifiers for this specific task deserves an in-depth exploration.  

Most important, the theoretical framework of \alg{} is designed to be completely general, hence it might be extended to include different scoring functions, generated via any arbitrary set of independent algorithmic strategies. 
In this regard, ongoing efforts aim at integrating detectors processing hidden layer features with others processing the properties of the output, and which have already proven their effectiveness in adversarial detection (see, e.g. \cite{hendrycksBaselineDetectingMisclassified2017,liangEnhancingReliabilityOutofdistribution2018,rothOddsAreOdd2019}).
Similarly, one may explore the possibility of exploiting the information on activation paths and/or regions, as suggested in \cite{luSafetyNetDetectingRejecting2017,craigheroInvestigatingCompositionalStructure2020}, as well as of refining the score definition by focusing on class-conditioned features. 

We finally note that the algorithmic framework of \alg{} was purposely designed to be as simple as possible, so to unequivocally show the advantages of adopting ensemble approaches in the “cleanest” scenario. 
However, improvements of the approach are expected to have important implications and should be pursued. 

First, the computational time of \alg{} is approximately the sum of that of the detectors employed to compute the layer-specific scores and, in the current version, it is mostly affected by OCSVM (see the computation time assessment in Supplementary Figure~2). 
Efforts should be devoted to tackle the trade-off between effectiveness and scalability, by adopting more efficient implementation schemes and/or constituting detectors. 

Second, effective strategies for the integration of the scoring functions might be devised.  As shown above, the effectiveness of the detectors and the importance of the layers is closely related to the attack type and, in our case, results in different optimal weights of the logit.  As a consequence, in case of unknown attacks, the selection of  weights may be complex or even unfeasible, impacting the detector performance. 
Alternative strategies to combine scores and/or detectors might be considered to improve the generability of our approach, e.g. via weighted averaging or voting schemes \cite{zhouEnsembleMethodsFoundations2012} or by employing test statistics \cite{raghuramGeneralFrameworkDetecting2021}, as well as via the exploitation of more robust feature selection and classification strategies.

Last, adaptive attacks \cite{carliniAdversarialExamplesAre2017}, i.e. adversarial example crafted to fool both the classifier and the detector, should be considered to further improve the design of the framework. Even though we expect that the heterogeneous nature of the ensemble would make it difficult to simultaneously fool all constituting detectors, an investigation of score aggregation schemes may be opportune to implement appropriate defense schemes against this kind of attacks.

To conclude, given the virtually limitless possibility of algorithmic extensions of our framework, the superior performance exhibited in benchmark settings in a purposely simple implementation, and the theoretical and application expected impact, we advocate a widespread and timely adoption of ensemble approaches in the field of adversarial detection.

\section*{Methods}
\label{sec:methods}

In this section, we illustrate the \alg{} ensemble approach for adversarial  detection, as well as the properties of the scoring functions it integrates, respectively based on OCSVM (Detector A - \emph{new}), Mahalanobis (Detector B) and LID (Detector C), which can also be used as stand-alone detectors. We also describe the background of both adversarial examples generation and detection, the partitioning of the input data and the extraction of the features processed by \alg{} and stand-alone detectors. 

\subsection*{Background}

\subsubsection*{Notation} 

Let us consider a multiclass classification problem with $C > 2$ classes. Let $\mathcal N(\vb*{x}) = \logit \circ h_L \circ \ldots \circ h_1(\vb*{x})$ be a DNN with $L$ layers, where $h_l$ is the $l^\text{th}$ hidden layer and $\logit$ is the output layer, i.e. the $C$ logits. To  simplify the notation, we refer to the activation of the  $l^\text{th}$ hidden layer given the input example $\vb*{x}$, i.e. to $h_l \circ \ldots \circ h_1(\vb*{x})$, as $h_l(\vb*{x})$ and to the logits as  $\logit(\vb*{x})$. 
Let $\softmax(\vb*{x}) = \softmax(\logit(\vb*{x}))$ 
be the softmax of $\logit(x)$, then the predicted class is $\hat y = \argmax_k \softmax(\vb*{x})^k$ with confidence $\hat p = \max_k \softmax(\vb*{x})^k$. Lastly, by $J(\vb*{x}, t) = - \log \softmax(\vb*{x})^t$ we will denote the cross-entropy loss function given input $\vb*{x}$ and target class~$t$.

\subsubsection*{Adversarial Attacks Algorithms}

In the following, we describe the adversarial attacks that were employed in our experiments, namely FGSM \cite{goodfellowExplainingHarnessingAdversarial2015}, BIM \cite{kurakinAdversarialExamplesPhysical2017}, DeepFool \cite{moosavi-dezfooliDeepFoolSimpleAccurate2016} and CW \cite{carliniEvaluatingRobustnessNeural2017}:

\begin{itemize}
\item FGSM \cite{goodfellowExplainingHarnessingAdversarial2015} defines optimal $L_\infty$ constrained perturbations as:
\[ 
\vb*{\tilde x} = \vb*{x} + \epsilon \cdot \sign{(\grad_{\vb*{x}}{J( \vb*{x}, t ) })},
\]
such that $\epsilon$ is the minimal perturbation in the direction of the gradient that changes the prediction of the model from the true class $y$ to the target class $t$.

\item BIM \cite{kurakinAdversarialExamplesPhysical2017} extends FGSM by applying it $k$ times with a fixed step size $\alpha$, while also ensuring that each perturbation remains in the $\epsilon$-neighbourhood of the original image $x$ by using a per-pixel clipping function $\clip$:
\begin{align*}
&\vb*{\tilde x_0} = \vb*{x} \\
&\vb*{\tilde x_{n+1}} = \clip_{\vb*{x}, \epsilon} {(\vb*{x_{n}} + \alpha \cdot \sign{(\grad_{\vb*{\tilde x_{n}}}{J( \vb*{\tilde x_{n}}, t ) })})}
\end{align*}

\item DeepFool \cite{moosavi-dezfooliDeepFoolSimpleAccurate2016} iteratively finds the optimal $L_2$ perturbations that are sufficient to change the target class by approximating the original non-linear classifier with a linear one. Thanks to the linearization, in the binary classification setting the optimal perturbation corresponds to the distance to the (approximated) separating hyperplane, while in the multiclass the same idea is extended to a one-vs-all scheme. In practice, at each step $i$ the method computes the optimal perturbation $p_i$ of the simplified problem, until $\vb*{\tilde x} = \vb*{x} + \sum_i p_i$ is misclassified.

\item the CW $L_2$ attack \cite{carliniEvaluatingRobustnessNeural2017} uses gradient descent to minimize $\norm{\vb*{\tilde x}-\vb*{x}}^2 + c \cdot l( \vb*{\tilde x} )$, where the loss $l$ is defined as:
\[ l(\vb*{x}) = \max{(\max \{\logit(\vb*{\tilde x})^i : i \neq t\}  - \logit(\vb*{\tilde x})^t, - \kappa )}. \]
The objective of the optimization is to minimize the $L_2$ norm of the perturbation and to maximize the difference between the target logit $\logit(\vb*{\tilde x})^t$ and the one of the next most likely class up to real valued constant $\kappa$, that models the desired confidence of the crafted adversarial.

\end{itemize}

\subsubsection*{Adversarial Detection}
\label{sec:detectors_sota}

Let us consider a classifier $\mathcal N$ trained on a training set $\mathcal X^{train}$ and a test example $\vb*{x_0} \in \mathcal X^{test}$, with predicted label $\hat y_0$. 
Adversarial examples detectors can be broadly categorized according to the features they consider: $(i)$ the features of the test example $\vb*{x_0}$, $(ii)$ the hidden features $h_l(\vb*{x_0})$, or $(iii)$ the features of the output of the network $out(\vb*{x_0})$, i.e., the logits $\logit(\vb*{x_0})$ or the confidence scores $\softmax(\vb*{x_0})$.

In brief, the first family of detectors tries to distinguish normal from adversarial examples by focusing on the input data. For instance, in \cite{hendrycksEarlyMethodsDetecting2017} the coefficients of low-ranked principal components of $\vb*{x_0}$ are used as features for the detector. In \cite{grosseStatisticalDetectionAdversarial2017}, the authors employed the statistical divergence, such as Maximum Mean Discrepancy, between $\mathcal X^{train}$ and $\mathcal X^{test}$ to detect the presence of adversarial examples in $\mathcal X^{test}$.

The second family of detectors aims at exploiting the information of the hidden features $h_l(\vb*{x_0})$.
In this group, some detectors rely on the identification of the nearest neighbors of $h_l(\vb*{x_0})$ to detect adversarial examples, by considering either: the conformity of the predicted class among the neighbors \cite{papernotDeepKNearestNeighbors2018, raghuramGeneralFrameworkDetecting2021}, the Local Intrinsic Dimensionality \cite{maCharacterizingAdversarialSubspaces2018}, the impact of the nearest neighbors on the classifier decision \cite{cohenDetectingAdversarialSamples2020} or the prediction of a graph neural-network trained on the nearest neighbors graph \cite{abusnainaAdversarialExampleDetection2021}.
In the same category, additional detectors take into account the conformity of the hidden representation $h_l(\vb*{x_0})$ to the hidden representation of instances with the same label in the training set, i.e. to $\{h_l(\vb*{x_0}) : \hat y_0 = y,  (\vb*{x}, y) \in X^{train}\}$, by computing either the Mahalanobis distance \cite{leeSimpleUnifiedFramework2018, kamoiWhyMahalanobisDistance2020} to the class means or the likelihoood of a Gaussian Mixture Model (GMM) \cite{zhengRobustDetectionAdversarial2018}.
Other detectors take the hidden representation $h_l(\vb*{x_0})$ itself as a discriminating feature, by training either a DNN \cite{metzenDetectingAdversarialPerturbations2017}, a Support Vector Machine (SVM) \cite{ luSafetyNetDetectingRejecting2017}, a One-class Support Vector Machine (OSVM) \cite{maNICDetectingAdversarial2019} or by using a kernel density estimate \cite{feinmanDetectingAdversarialSamples2017}. Additional methods within this family use the hidden representation $h_l(\vb*{x_0})$ as feature to train a predictive model $m_l$. The model $m_l$ either seeks to predict the same $C$ classes of the original classifier \cite{maNICDetectingAdversarial2019, sotgiuDeepNeuralRejection2020} or to reconstruct the input data $\vb*{x_0}$ from $h_l(\vb*{x_0})$ \cite{hendrycksEarlyMethodsDetecting2017, hendrycksBaselineDetectingMisclassified2017}.
The detector then classifies $\vb*{x_0}$ as adversarial/benign by relying on the confidence of the prediction $\hat{y}_0$ in the former case, on its reconstruction error in the latter.

The third family of detectors employs the output of the network $out(\vb*{x_0})$ to detect adversarial inputs. In this category, some detectors considers the divergence between $out(\vb*{x_0})$ and $out(\phi(\vb*{x_0}))$ where $\phi$ is a function such as a squeezing function that reduces the features of the input \cite{xuFeatureSqueezingDetecting2018}, an autoencoder trained on $\mathcal X^{train}$ \cite{mengMagNetTwoprongedDefense2017}, a denoising filter \cite{liangDetectingAdversarialImage2021} or a random perturbation \cite{rothOddsAreOdd2019, huangModelagnosticAdversarialDetection2019}. Some others take the confidence score of the predicted class $\hat y$, that is expected to be lower when the example is anomalous \cite{hendrycksEarlyMethodsDetecting2017, hendrycksBaselineDetectingMisclassified2017, liangEnhancingReliabilityOutofdistribution2018}. Lastly, in \cite{feinmanDetectingAdversarialSamples2017} Bayesian uncertainty of dropout DNNs was used as a feature for the detector.

\subsection*{Data partitioning}
\label{sec:data_partitioning}
Let $\mathcal X^{train}$ be the training set on which the classifier $\mathcal N$ was trained, $\mathcal X^{test}$ the test set and $\mathcal L_{norm}\subseteq{\mathcal X^{test}}$ the set of correctly classified test instances. Following the setup done in \cite{maCharacterizingAdversarialSubspaces2018, leeSimpleUnifiedFramework2018}, from $\mathcal L_{norm}$ we generate $(i)$ a set of noisy examples $\mathcal L_{noisy}$ by adding random Gaussian noise, with the additional constraint of being correctly classified, and $(ii)$ a set of adversarial examples $\mathcal L_{adv}$ generated via a given attack. We also ensure that $\mathcal L_{norm}$, $\mathcal L_{noisy}$ and $\mathcal L_{adv}$ have the same size.
The set $\mathcal L = \mathcal L_{norm} \cup \mathcal L_{noisy} \cup \mathcal L_{adv}$ will be our \emph{labelled dataset}, where the label is $adv$ for adversarial examples and $\overline{adv}$ for benign ones. As detailed in the following sections, $\mathcal L$ will be split into a training set $\mathcal L^{train}$, a validation set $\mathcal L^{valid}$ for hyperparameter tuning and a test set $\mathcal L^{test}$ for the final evaluation.

\subsection*{Feature Extraction}
\label{sect:feat_extract}

In our experimental setting, $h_l(\vb*{x})$, with $l \in [1, \dots, L]$, corresponds to either the first convolutional layer or to the output of the $l^{th}$ dense (residual) block of a DNN (e.g. DenseNet or ResNet). 
As proposed in \cite{leeSimpleUnifiedFramework2018}, the size of the feature map is reduced via average pooling, so that $h_l(\vb*{x})$ has a number of features equal to the number of channels of the $l^{th}$ layer.  
Detectors A, B, and C and \alg{} are applied to such set of features, as detailed in the following. 

\subsection*{Detector A: OCSVM}
\label{sec:ocsvm}

\begin{algorithm}[t]
\caption{OCSVM detector (see the main text for an explanation of the notation employed) }
\label{alg:ocsvm}
\begin{algorithmic}[1]
\Require Act. $h_l$ of layer $l$, trainset $\mathcal X^{train}$, labelled set $\mathcal L$
\ForEach{$l$ in $1, \dots, L$}
    \State Centering and PCA-whitening of $h_l$: $h_l^*$
    \State Select best layer-specific parameters $\theta = \{ \nu, \gamma \}$
    \State Fit OCSVM$_l(\theta)$ on $\{ h_l^*(\vb*{x}) : \vb*{x} \in \mathcal X^{train} \}$
    \State OCSVM$_l(\theta)$ decision function: $\text{O}_l$
    \State Layer $l$ score of $\vb*{x_0}$: $\text{O}_l(\vb*{x_0})$
\EndFor
\State Scores vector: $\textbf{O}(\vb*{x_0}) := [\text{O}_1(\vb*{x_0}), \dots, \text{O}_L(\vb*{x_0})]$
\State Fit $adv$ posterior on $\mathcal L^{train}$: $p( adv \rvert \textbf{O}(\vb*{x_0}))$
\State \textsf{OCSVM} of $\vb*{x_0}$: \textsf{OCSVM}$(\vb*{x_0}) \coloneqq p( adv \rvert \textbf{O}(\vb*{x_0}))$ \\
\Return \textsf{OCSVM}
\end{algorithmic}
\end{algorithm}

This newly designed detector is based on a standard anomaly detection technique called One-Class SVM (OCSVM) \cite{scholkopfSupportVectorMethod1999}, which belongs to the family of one-class classifiers \cite{taxOneclassClassificationConceptlearning2001}. One-class classification is a problem in which the classifier aims at learning a good description of the training set and then rejects the inputs that do not resemble the data it was trained on, which represent outliers or anomalies. 
This kind of classifiers is usually adopted when only one class is sufficiently represented within the training set, while the others are undersampled or hard to be characterized, as in the case of adversarial examples, or anomalies in general. 
OCSVM was first employed for adversarial detection in \cite{maNICDetectingAdversarial2019}. Here, we modified it by defining an input pre-processing step based on PCA-whitening \cite{kessyOptimalWhiteningDecorrelation2018}, and by employing a Bayesian optimization technique \cite{snoekPracticalBayesianOptimization2012} for hyperparameter tuning.
The pseudocode is reported in Algorithm~\ref{alg:ocsvm}.

\subsubsection*{Preprocessing}
OCSVM employs a kernel function (in our case a Gaussian RBF kernel) that computes the Euclidean distance among data points. Hence, it might be sound to standardize all the features of the data points, at the preprocessing stage, to make them equally important. 
To this end, each hidden layer activation $h_l(\vb*{x_0})$  is first centered on the mean activations $\mu_{ l, c}$ of the examples of the training set $\mathcal X^{train}$ of class $c$. Then, PCA-whitening $\vb{W}^{PCA}_l$ is applied: 
\begin{align*}
    h_l^*(\vb*{x_0}) &= \vb{W}^{PCA}_l \cdot (h_l(\vb*{x_0}) - \mu_{l, c})\\
    &= \vb*{\Lambda}^{-1/2}_l \cdot \vb*{U}^T_l \cdot (h_l(\vb*{x_0}) - \mu_{l, c}),
\end{align*} 
where $\vb*{U}^T_l$ is the eigenmatrix of the covariance $\vb{\Sigma}_l$ of activations $h_l$ and $\vb*{\Lambda}_l$ is the eigenvalues matrix of the examples of $\mathcal X^{train}$. Whitening is a commonly used preprocessing technique for outlier detection, since it enhances the separation of points that deviate in low-variance directions \cite{aggarwalLinearAlgebraOptimization2020}. 
Moreover, in \cite{kamoiWhyMahalanobisDistance2020} it was conjectured that the effectiveness of the Mahalanobis distance \cite{leeSimpleUnifiedFramework2018} for out-of-distribution and adversarial detection is due to the strong contribution of low-variance directions. Thus, this preprocessing step allows the one-class classifier to achieve better overall performances\footnote{In a test on the DenseNet, CIFAR-10, CW scenario, the AUROC returned by the OCSVM detector with PCA-whitening preprocessing improves from $82.56$ to $90.24$, and the AUPR from $78.17$ to $82.98$, with respect to the same method without preprocessing (see Sect.~\ref{sec:results} for further details).}. 

\subsubsection*{Layer-specific scoring function}

After preprocessing, a OCSVM with a Gaussian RBF kernel is trained on the   hidden layer activations $h_l$ of layer  $l$ of the training set $\mathcal X^{train}$. 
Once the model has been fitted, for each instance $\vb*{x_0}$ layer-specific scores  $\text{\textbf{O}}(\vb*{x_0}) = [\text{O}_1(\vb*{x_0}), \text{O}_2(\vb*{x_0}), \dots, \text{O}_L(\vb*{x_0})]$ are evaluated.
More in detail, let $\mathcal S_l$ be the set of support vectors, the decision function $\text O_l(\vb*{x_0})$ for the $l^{\text{th}}$ layer is computed as: 
\begin{equation}
\text{O}_l(\vb*{x_0}) = \sum_{sv \in \mathcal S_l} \alpha_{sv} k(h_l(\vb*{x_0}), sv) - \rho,
\end{equation}
where $\alpha_{sv}$ is the coefficient of the support vector $sv$ in the decision function, $\rho$ is the intercept of the decision function and $k$ is a Gaussian RBF kernel with kernel width $\gamma$:
\begin{equation}
k(\vb*{x}, \vb*{y}) = \exp (- \gamma \norm{\vb*{x} - \vb*{y}}^2 ).
\end{equation}

\subsubsection*{Hyperparameter optimization} 
\begin{figure}[t]
\begin{center}
\includegraphics[width=\linewidth/2]{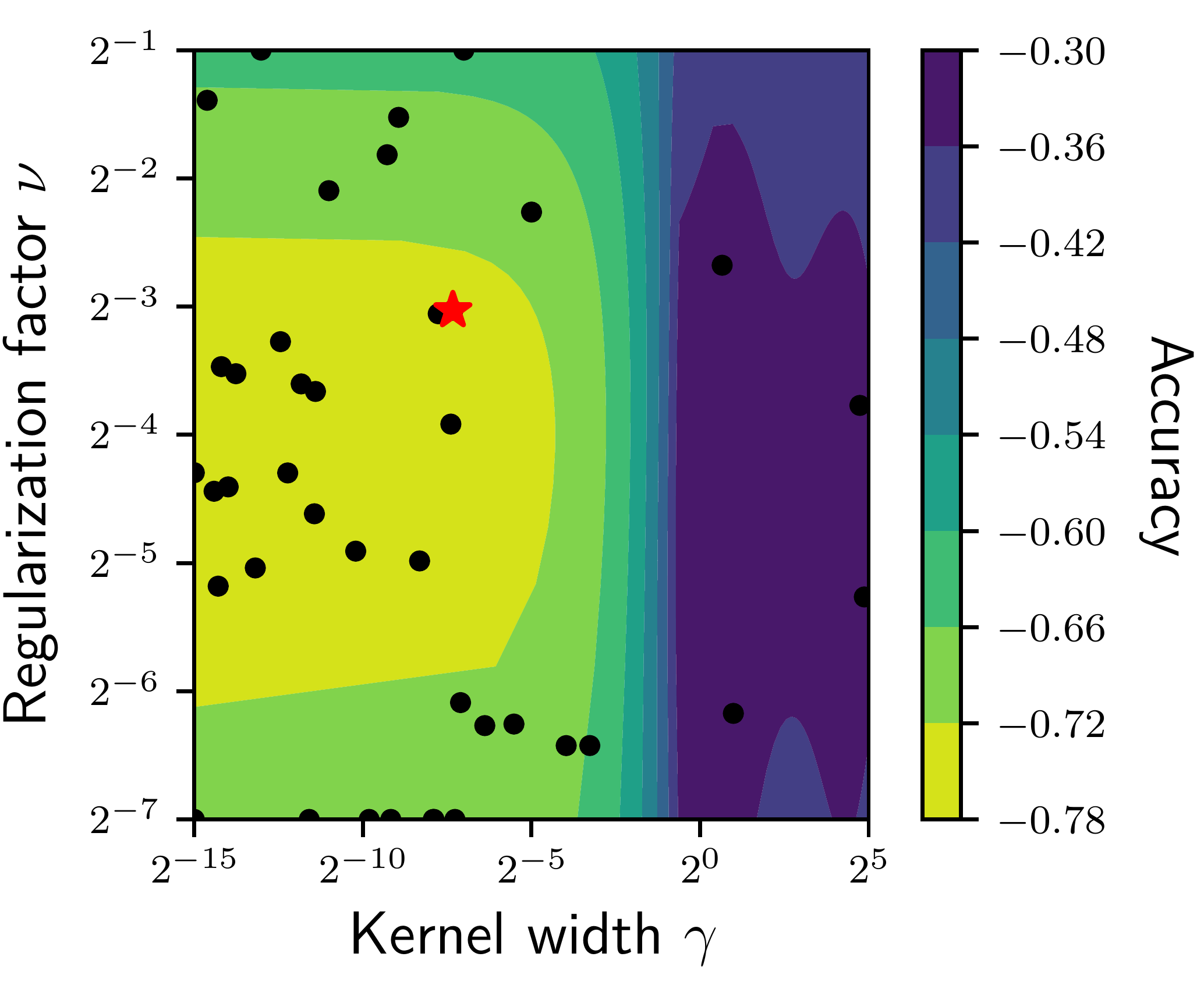}
\end{center}
\caption{{\bfseries OCSVM hyperparameter optimization.} The influence of different combinations of OCSVM hyperparameters $\{ \nu, \gamma \}$ on the validation accuracy is explored via Bayesian optimization \cite{headScikitoptimizeSequentialModelbased}, in the example scenario of DenseNet model, CIFAR-10 dataset and DeepFool attack. The gradient returns the validation accuracy estimated on $\mathcal L^{valid}$.  The red star represents the optimal configuration, which is then employed for adversarial detection in both the OCSVM and the \alg{} detectors.}
\label{fig:paramscan}
\end{figure}

The layer-specific scoring function takes two parameters as input: the regularization factor $\nu \in (0,1)$ that controls the fraction of training errors that should be ignored, and the kernel width $\gamma$. This hyperparameters must be carefully chosen to achieve good performances. 
For this purpose, many approaches have been proposed for hyperparameters selection in OCSVMs \cite{alamOneclassSupportVector2020}. In our setting, we used the validation set of labelled examples $\mathcal L^{valid}$ to choose the best combination of parameters, based on the validation accuracy. To avoid a full (and infeasible) exploration of the parameters space, we employed Bayesian hyperparameter optimization, via the \texttt{scikit-optimize} library \cite{headScikitoptimizeSequentialModelbased}. In Figure~\ref{fig:paramscan}, we report the estimated accuracy of the explored solutions in the specific case of the OCSVM detector, in a representative experimental setting.

\subsubsection*{Adversarial detection}
Once the scores have been obtained for each layer, they can serve either as input for the stand-alone detector A or as partial input of the ensemble detector EAD.
In the former case, in order to aggregate the scores of the separate layers $\text{\textbf{O}}(\vb*{x_0})$,  this detector employs a logistic regression to model the posterior probability of adversarial ($adv$) examples:
\begin{equation}
\begin{split}
    & p( adv \rvert \text{\textbf{O}}(\vb*{x_0})) = \Big(1 + \exp(\beta_0 + \vb*{\beta}^T \text{\textbf{O}}(\vb*{x_0})) \Big)^{-1}, \\
\end{split}
\end{equation}
The parameters $\{\beta_0, \vb*{\beta} \}$ are fitted with a cross-validated procedure using the labelled training set $\mathcal L^{train}$.

\subsection*{Detector B: Mahalanobis}
\label{sec:maha}

The Mahalanobis detector was originally introduced in \cite{leeSimpleUnifiedFramework2018}. The algorithmic procedure is akin to that of the OCSVM detector, and includes a final layer score aggregation step via logistic regression, but it is based on a different layer-specific scoring function.

\subsubsection*{Layer-specific scoring function}

Given a test instance $\vb*{x_0}$, the layer score  is computed via a three-step procedure: first, for each instance,  the class  $\hat c$  is selected, such that:
\[ \hat c = \argmin_c \text{Maha}_l(\vb*{x_0}, c),\]
where $\text{Maha}_l(\vb*{x}, c)$ is the Mahalanobis distance for the $l^{\text{th}}$ layer between the activations $h_l(\vb*{x})$ and the mean values $\mu_{l,c}$ of the examples in the training set $\mathcal X^{train}$:
\begin{equation}
    \text{Maha}_l(\vb*{x}, c) = (h_l(\vb*{x}) - \mu_{l, c})^T \vb*{\Sigma}_l^{-1} (h_l(\vb*{x}) - \mu_{l, c}),
\end{equation}
where $\vb*{\Sigma}_l$ is the covariance matrix of the examples of $\mathcal X^{train}$ in layer $l$.
Then, the instance is preprocessed to obtain a better separation between benign and adversarial examples similarly to what discussed in \cite{liangEnhancingReliabilityOutofdistribution2018}:

\[ \vb*{x_0^*} = \vb*{x_0} - \lambda \sign{\grad_{\vb*{x_0}} \text{Maha}_l(\vb*{x_0}, \hat c)}, \]
where $\lambda$ is positive real number, called the perturbation magnitude.
The scoring for instance $\vb*{x_0}$ is computed as:
\[ \text{\textbf{M}}(\vb*{x_0}) = [\text{M}_1(\vb*{x_0}), \text{M}_2(\vb*{x_0}), \dots, \text{M}_L(\vb*{x_0})], \]
where 
\[ \text{M}_l(\vb*{x_0}) = - \max_c \ {\text{Maha}_l(\vb*{x_0^*}, c)}.\]

\subsubsection*{Adversarial detection}
The scores can serve either as input for the stand-alone detector B or as partial input of the ensemble detector EAD.
In the latter case, the Mahalanobis detector uses a logistic regression to identify adversarial examples, with a procedure  similar to that already described for stand-alone detector A.

\subsubsection*{Hyperparameter optimization} 
Differently from detector A, the hyperparameter selection is performed downstream of the adversarial detection stage. In order to select the best $\lambda$ (unique for all layers), the method selects the value that achieves the best Area Under the Receiver Operating Characteristic (AUROC) on $\mathcal L^{valid}$ computed on the posterior probability $p(adv \rvert \text{\textbf{M}}(\vb*{x_0}))$, which is obtained via the logistic regression fitted on $\mathcal L^{train}$ (the definition of AUROC is provided in the Supplementary Material).

\subsection*{Detector C: LID}
\label{sec:lid}

The third detector uses a procedure similar to detectors A-B, but the layer-specific scoring function is based on the Local Intrinsic Dimensionality (LID) approach \cite{maCharacterizingAdversarialSubspaces2018}. 

\subsubsection*{Layer-specific scoring function}

Given a test instance $\vb*{x_0}$, the LID layer-specific scoring function L is defined as:

\begin{equation}
    \text{L}_l(\vb*{x_0}) = - \bigg(\frac{1}{k} \sum_{i=1}^k \log{\frac{r_i(h_l(\vb*{x_0}))}{\max_i r_i(h_l(\vb*{x_0}))}} \bigg)^{-1},
\end{equation}
where,  $k$ is the number of nearest neighbors, $r_i$ is the Euclidean distance to the $i$-th nearest neighbor in the set of normal examples $\mathcal L_{norm}$. The layer-specific scores are:
\[ \text{\textbf{L}}(\vb*{x_0}) = [\text{L}_1(\vb*{x_0}), \text{L}_2(\vb*{x_0}), \dots, \text{L}_L(\vb*{x_0})] \]

\subsubsection*{Adversarial detection}
When considered alone, the LID detector employs a logistic regression to identify adversarial examples, similarly to the other detectors (see above). 

\subsubsection*{Hyperparameter optimization}  Similarly to detector B, the hyperparameter selection is performed downstream of the adversarial detection stage.  $k$ is selected as the value that achieves the best AUROC on $\mathcal L^{valid}$ computed on the posterior probability $p(adv \rvert \text{\textbf{L}}(\vb*{x_0}))$, which is obtained via the logistic regression fitted on $\mathcal L^{train}$. Note that $k$ is unique for all  layers.

\subsection*{ENsemble Adversarial Detector (\alg{})}
\label{sec:enad}

The \alg{} approach exploits the effectiveness of detectors A, B, and C in capturing different properties of data distributions, by explicitly integrating the distinct layer-specific scoring functions in a unique classification framework.
More in detail, given a test instance $\vb*{x_0}$,  it will be characterized by a set of layer-specific and detector-specific features, computed from the scoring functions defined above, that is: $\textbf{E}(\vb*{x_0}) = [\textbf{O}(\vb*{x_0}), \textbf{M}(\vb*{x_0}), \textbf{L}(\vb*{x_0})]$. 
It should be noted that training and hyperparameter optimization is executed for each detector independently. 

\subsubsection*{Adversarial detection}
In its current implementation, in order to integrate the scores of the separate layers $\textbf{E}(\vb*{x_0})$, \alg{} employs a simple logistic regression to model the posterior probability of adversarial ($adv$) examples:
\begin{equation}
\begin{split}
    & p( adv \rvert \textbf{E}(\vb*{x_0})) = \Big(1 + \exp(\beta_0 + \vb*{\beta}^T \textbf{E}(\vb*{x_0})) \Big)^{-1}. \\
\end{split}
\end{equation}

Like detectors A, B, and C, the logistic is fitted with a cross-validation procedure using the labelled training set $\mathcal L^{train}$. Fitting the logistic allows one to have different weights, i.e. the elements of $\vb*{\beta}^T$, for the different layers and detectors, meaning that a given detector might be more effective in isolating an adversarial example when processing its activation on a certain layer of the network. The pseudocode is reported in Algo.~\ref{alg:ead}.

\begin{algorithm}[t]
\caption{\alg{} detector.}
\label{alg:ead}
\begin{algorithmic}[1]
\Require Act. $h_l$ of layer $l$, trainset $\mathcal X^{train}$, labelled set $\mathcal L$

\State Select best hyperparameters for OCSVM, Maha, LID
\ForEach{layer $l$ in $1, \dots, L$}
    \State Layer $l$ scores of $\vb*{x_0}$: $\text{O}_l(\vb*{x_0}), \text{M}_l(\vb*{x_0}), \text{L}_l(\vb*{x_0})$
\EndFor
\State Scores vector: $\textbf{E}(\vb*{x_0}) \coloneqq [\textbf{O}(\vb*{x_0}), \textbf{M}(\vb*{x_0}), \textbf{L}(\vb*{x_0})]$
\State Fit $adv$ posterior on $\mathcal L^{train}$: $p( adv \rvert\textbf{E}(\vb*{x_0}))$
\State \alg{} on $\vb*{x_0}$: \alg{}$(\vb*{x_0}) \coloneqq p( adv \rvert \textbf{E}(\vb*{x_0}) )$ \\
\Return \alg{}
\end{algorithmic}
\end{algorithm}

\subsection*{Performance Metrics}
\label{sec:perf_metrics}

Let the positive class be the adversarial examples ($adv$) and the negative class be the benign examples ($\overline{adv}$). Then, the correctly classified adversarial and benign examples correspond to the true positives ($\text{TP}$) and true negatives ($\text{TN}$), respectively. Conversely, the wrongly classified adversarial and benign examples are the false negatives ($\text{FN}$) and false positives ($\text{FP}$), respectively.

To evaluate the detectors performances, we employed two standard threshold independent metrics, namely \textsf{AUROC} and \textsf{AUPR} \cite{davisRelationshipPrecisionrecallROC2006}, and the accuracy, defined as follows:

\begin{itemize}

    \item Area Under the Receiver Operating Characteristic curve (\textsf{AUROC}): the area under the curve identified by $\textsf{specificity} = \text{TN} /(\text{TN} + \text{FP})$ and $\textsf{fall-out} = \text{FP} /(\text{FP} + \text{TN})$.
    
    \item Area Under the Precision-Recall
curve (\textsf{AUPR}): the area under the curve identified by $\textsf{precision} = \text{TP} /(\text{TP} + \text{FP})$ and  $\textsf{recall} = \text{TP} /(\text{TP} + \text{FN})$.

    \item $\textsf{Accuracy} = (\text{TP}+\text{TN})/(\text{TP} + \text{TN} +\text{FP} +\text{FN})$.
\end{itemize}

The \textsf{AUROC} and \textsf{AUPR} were evaluated given the adversarial posterior learned by the logistic function $p( adv \rvert  X )$, were $X$ is the set of layer-specific scores. The layer-specific scores are computed from $\mathcal L^{valid}$ when the AUROC is used for hyperparameters optimization for the Mahalanobis and LID detectors, and from $\mathcal L^{test}$ in all the other settings, i.e. the detectors performance evaluation. 
Moreover, \textsf{AUROC} and \textsf{AUPR} were also used to evaluate the performance of each detector in each layer, by consider the layer-specific scores evaluated on $\mathcal L^{test}$ (see,. e.g., Figure~\ref{fig:layer-imp}). Note that for the OCSVM and Mahalanobis detectors, lower values corresponds to adversarial examples, while the opposite applies for the LID detector. 
Finally, the accuracy was used for the Bayesian hyperparameter selection procedure \cite{snoekPracticalBayesianOptimization2012} of the OCSVM detector.

\printbibliography

\end{document}


\title{[Supplementary]\\Unity is strength: improving the detection of adversarial examples with ensemble approaches}

\author[1]{Francesco Craighero}
\author[1]{Fabrizio Angaroni}
\author[1]{Fabio Stella}
\author[2,3]{Chiara Damiani}
\author[1,3,4]{Marco Antoniotti}
\author[5,1,3]{Alex Graudenzi}

\affil[1]{Dept. of Informatics, Systems and Communication, Universit\`{a} degli Studi di Milano-Bicocca, Milan, Italy}

\affil[2]{Dept. of Biotechnology and Biosciences, Universit\`{a} degli Studi di Milano-Bicocca, Milan, Italy}

\affil[3]{B4 - Bicocca Bioinformatics Biostatistics and Bioimaging Centre, Universit\`{a} degli Studi di Milano-Bicocca, Milan, Italy}

\affil[4]{Bioinformatics Program, Tandon School of Engineering, NYU Poly, New York, NY, U.S.A}

\affil[5]{Institute of Molecular Bioimaging and Physiology, National Research Council (IBFM-CNR), Milan, Italy}

\affil[ ]{\email{f.craighero@campus.uinimib.it}, \email{fabrizio.angaroni@unimib.it}, \email{fabio.stella@unimib.it}, \email{chiara.damiani@unimib.it}.
\email{marco.antoniotti@unimib.it}, \email{alex.graudenzi@unimib.it}}

\date{}

\maketitle

\section{Supplementary Results}

\subsection{Selected hyperparameters}
The hyperparameter optimization was executed separately for detectors A, B and C, by scanning the configurations reported in Table~\ref{supp_tab:paramscan}.
Let $\mathcal L$ be the set of unseen adversarial and benign examples which are, by definition, wrongly and correctly classified, respectively. In all tests, the hyperparameter optimization was performed on the validation set $\mathcal L^{valid}$, whereas the logistic regression fit on the training set $\mathcal L^{train}$ of the same attack, except for the unknown attack scenario, in which both the hyperparameters and the logit weights were estimated on the FGSM attack and employed with the others. 

The selected hyperparameters for each experimental setting are shown in Supplementary Tables \ref{supp_tab:ocsvm_best_hyp} and \ref{supp_tab:maha_lid_best_hyp}. 

\begin{table}
    \begin{center}
    \begin{tabular}{ccc}
    \toprule
    Detector & Parameter & Configurations \\
    \midrule
    \multirow{2}{*}{OCSVM} & $\nu$ & $2^{-7}, 2^{-6}, \dots, 2^{-1}$ \\
    & $\gamma$ & $2^{-15}, 2^{-14}, \dots, 2^{5}$ \\
    \cmidrule{1-3}
    LID & $k$ & $10, 20, \dots, 90$ \\
    \cmidrule{1-3}
    Maha &  $\lambda$ & \begin{tabular}{@{}c@{}}$ 0.0, 0.01, 0.005, 0.002,$ \\ $0.0014, 0.001, 0.0005$\end{tabular} \\
    \bottomrule
    \end{tabular}
    \end{center}
    \caption{{\bfseries Hyperparameters configurations.} Hyperparameters space explored in the optimization step for the three detectors OCSVM, LID and Mahalanobis. }
    \label{supp_tab:paramscan}
\end{table}

\subsection{Additional comparisons of stand-alone detectors}

\begin{figure}[t]
\begin{center}
\includegraphics{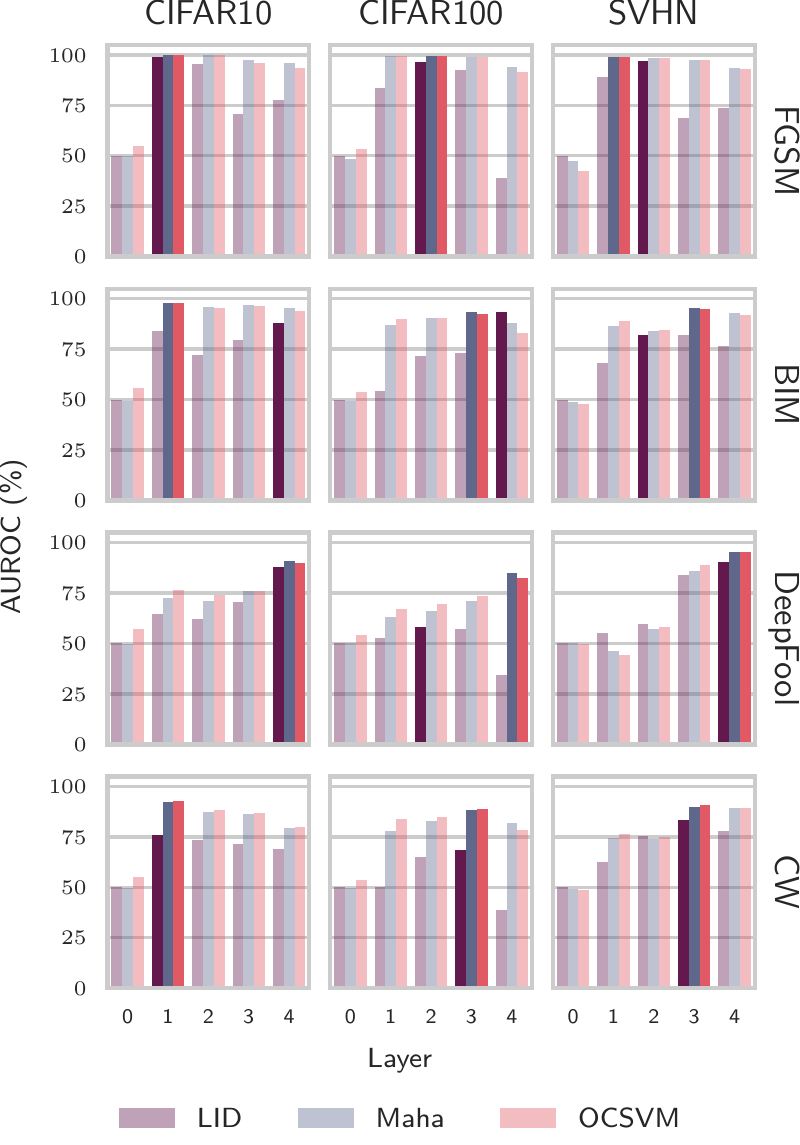}
\end{center}
\caption{{\bfseries Influence of layers in adversarial detection (ResNet model).} For each configuration of datasets and attacks on the ResNet model, the AUROC of each layer-specific score for OCSVM, LID and Mahalanobis detectors is returned. For each configuration and detector, the best performing layer is highlighted with a darker shade.}
\label{supp_fig:barplots}
\end{figure}

In Figure~\ref{supp_fig:barplots} the layer importance of the ResNet model in all the settings is illustrated. The layer importance is evaluated by the AUROC of the layer-specific scores, i.e. $\text{M}_l$, $\text{O}_l$ and $\text{L}_l$ for the Mahalanobis, OCSVM and LID detectors, respectively (see above). 
Interestingly, the most important layer is mostly dependant on the attack type, but also on the dataset and detector type. Indeed, in the DeepFool attack the most important layer is the last one, while in the other attacks the highest values of AUROC are mainly achieved in the second and fourth layers. Moreover, for CIFAR-10 the second layer is the most important in the BIM and CW settings, while in the other attacks the most important is the fourth. Last, in the BIM attack, the LID detector behaves differently than the other detectors in all settings. 

\begin{figure*}[hb]
\centering
\includegraphics{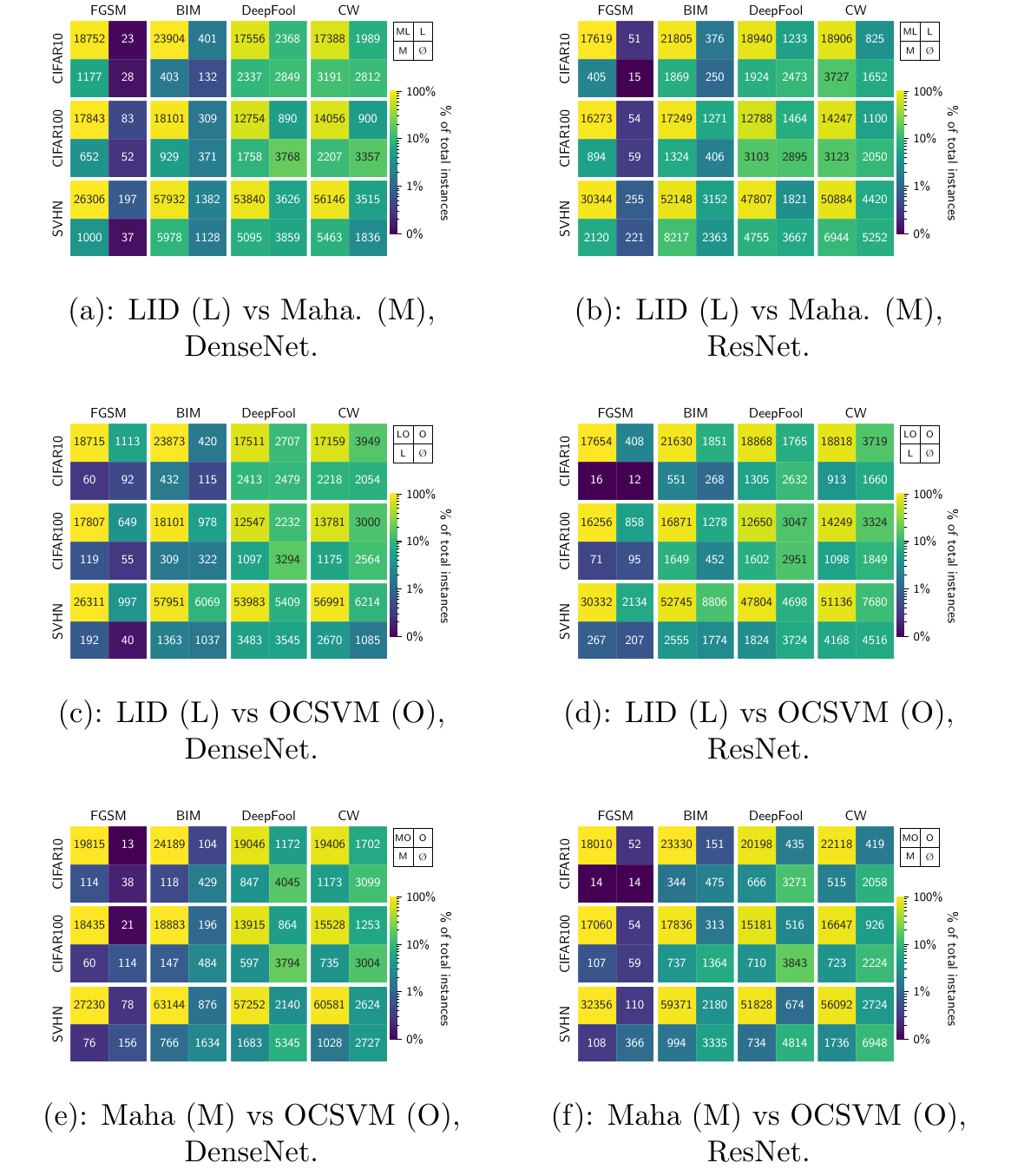}
\caption{{\bfseries Comparison of predictions of single detectors.} The contingency table shows the number of adversarial examples correctly identified: by both the detectors (top-left box), by either one of the two methods (diagonal boxes), by none of them (lower-right box), in all the experimental settings described in the main text. }
\label{supp_fig:other_heatmaps}
\end{figure*}

In Figure~\ref{supp_fig:other_heatmaps}, we report the pairwise comparisons of all detectors on all experimental settings as contingency tables (see Results for further details). The most meaningful observations can be obtained by looking at the diagonals of the table in each configuration, in which the instances detected by either one of the two detectors are counted. The greatest differences are observed in the case of DeepFool and CW attacks, especially with respect to the LID detector, e.g. 4168 vs 7680 instances in the SVHN, CW, ResNet setting of the LID/OCSVM comparison.

\begin{figure*}[hb]
    \centering
    \includegraphics[width=.75\linewidth]{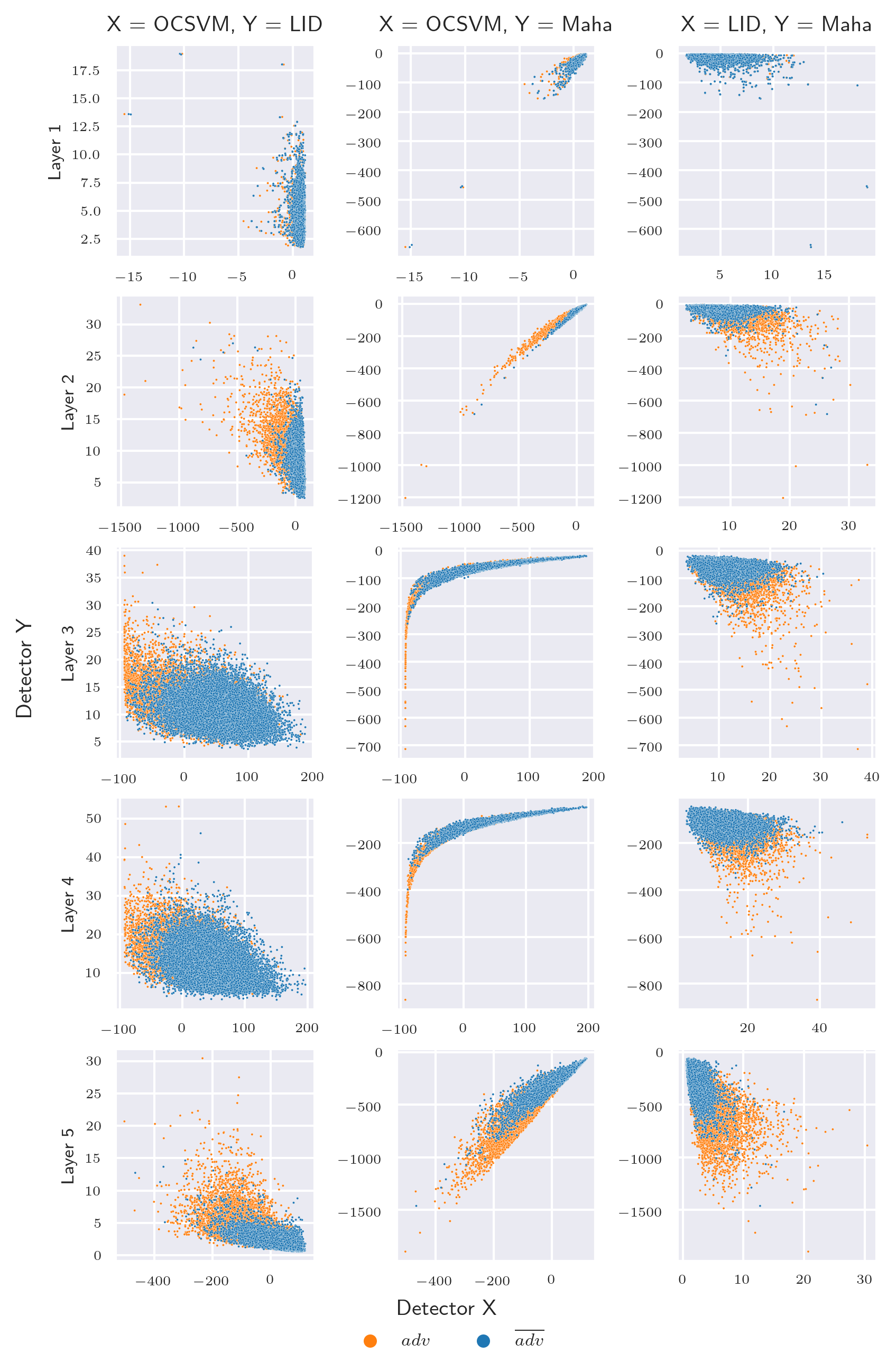}
    \caption{{\bf Pairwise layer-specific scores comparison.} Comparison of the layer scores of OCSVM, LID and Mahalanobis detectors, i.e. $\text{O}_l$, $\text{L}_l$ and $\text{M}_l$, respectively, for each layer $l$ and for the Resnet, CIFAR-10, DeepFool configuration. }
    \label{supp_fig:scatters}
\end{figure*}

In Figure~\ref{supp_fig:scatters}, we show the pairwise comparisons of the layer-specific scores, i.e. $\text{M}_l$, $\text{O}_l$ and $\text{L}_l$ for the Mahalanobis, OCSVM and LID detectors, respectively, in the ResNet, CIFAR-10, DeepFool setting. 
The most correlated detectors appear to be the OCSVM and Mahalanobis detectors (second column), mainly in layers 2 and 4 (second and fourth row). Such correlation is likely due to the pre-processing step employed in the OCSVM detector, i.e. PCA-whitening, which is closely related (by a rotation) to the ZCA-Mahalanobis whitening  that is used to compute the Mahalanobis distance \cite{kessyOptimalWhiteningDecorrelation2018}.

\subsection{Computation time}

All tests were executed on a \textsf{n1-standard-8} Google Cloud Platform instance, with eight quad-core Intel\textsuperscript{\textregistered} Xeon\textsuperscript{\textregistered} CPU (2.30GHz), 30GB of RAM and a NVIDIA Tesla\textsuperscript{\textregistered} K80.

\begin{figure}[t]
    \centering
    \includegraphics[scale=.7]{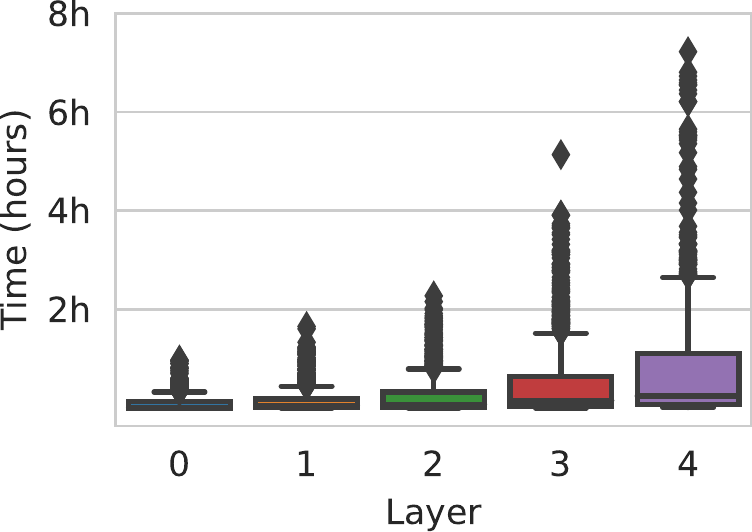}
    \caption{\textbf{OCSVM fit times by layer.} Fitting times of the OCSVM stand-alone detector, for each configuration reported in Table~\ref{supp_tab:ocsvm_best_hyp}. Attack, model and dataset are aggregated, since the mean fit time depends only on the layer on which the detector was trained, i.e. the deeper the layer, the higher the mean fit time.}
    \label{supp_fig:ocsvm_fit_times}
\end{figure}

The fitting time of each parameter explored in all configurations for OCSVM is reported in Figure~\ref{supp_fig:ocsvm_fit_times}. For the computation times of the Mahalanobis and LID detectors, please refer to \cite{leeSimpleUnifiedFramework2018} and \cite{maCharacterizingAdversarialSubspaces2018}.

\begin{sidewaystable*}[hb]
    \begin{center}
    \footnotesize
    \begin{tabular}{ccccccccccccc}
    \toprule
           &      & {} & \multicolumn{10}{c}{Best Value} \\
           &      & Layer & \multicolumn{2}{c}{0} & \multicolumn{2}{c}{1} & \multicolumn{2}{c}{2} & \multicolumn{2}{c}{3} & \multicolumn{2}{c}{4} \\
           &      & Parameter &      $\nu$ & $\gamma$ &    $\nu$ & $\gamma$ &    $\nu$ & $\gamma$ &    $\nu$ & $\gamma$ &    $\nu$ & $\gamma$ \\
    Model & Dataset & Attack &            &          &          &          &          &          &          &          &          &          \\
    \midrule
    \multirow{12}{*}{DenseNet} & \multirow{4}{*}{CIFAR-10} & FGSM &    1.27e-2 &  9.43e-2 &  7.81e-3 &  2.53e-4 &  7.81e-3 &  1.79e-4 &  2.06e-2 &  3.05e-5 &          &          \\
           &      & BIM &    9.40e-3 &  9.78e-3 &  4.15e-2 &  1.23e-2 &  1.87e-2 &  1.45e-2 &  1.26e-2 &  6.08e-5 &          &          \\
           &      & DeepFool &    1.12e-2 &  3.05e-5 &  7.81e-3 &  1.87e-2 &  1.29e-1 &  9.86e-5 &  1.23e-1 &  6.21e-3 &          &          \\
           &      & CW &    1.13e-2 &  1.32e-4 &  4.19e-2 &  2.86e-2 &  1.17e-1 &  1.65e-2 &  5.69e-2 &  8.62e-3 &          &          \\
    \cline{2-13}
           & \multirow{4}{*}{CIFAR-100} & FGSM &    8.25e-2 &  2.81e-4 &  7.81e-3 &  3.05e-5 &  7.81e-3 &  3.05e-5 &  2.04e-2 &  3.05e-5 &          &          \\
           &      & BIM &    1.41e-2 &  1.75e-2 &  7.81e-3 &   1.7e-2 &  6.17e-2 &  7.93e-4 &  2.09e-2 &  4.11e-3 &          &          \\
           &      & DeepFool &    1.10e-2 &   1.2e-4 &  8.81e-2 &  7.83e-3 &  4.67e-2 &  1.01e-2 &  1.08e-1 &  2.51e-4 &          &          \\
           &      & CW &    7.81e-3 &  3.05e-5 &  1.22e-1 &  9.69e-3 &  1.37e-1 &  3.35e-5 &  1.10e-1 &  3.05e-5 &          &          \\
    \cline{2-13}
           & \multirow{4}{*}{SVHN} & FGSM &    7.81e-3 &   6.1e-2 &  7.81e-3 &  2.52e-4 &  7.81e-3 &  4.88e-5 &  7.81e-3 &  3.05e-5 &          &          \\
           &      & BIM &    7.81e-3 &  2.32e-4 &  2.17e-2 &  2.98e-4 &  4.03e-2 &  7.63e-4 &  2.45e-2 &  3.05e-5 &          &          \\
           &      & DeepFool &    7.81e-3 &  3.05e-5 &  1.45e-2 &  1.74e-3 &  2.84e-2 &  3.05e-5 &  4.55e-2 &  4.16e-4 &          &          \\
           &      & CW &    7.81e-3 &  3.05e-5 &  1.18e-2 &  1.35e-2 &  8.92e-3 &  1.55e-2 &  1.43e-2 &  7.18e-3 &          &          \\
    \cline{1-13}
    \cline{2-13}
    \multirow{12}{*}{ResNet} & \multirow{4}{*}{CIFAR-10} & FGSM &    7.81e-3 &  3.05e-5 &  7.81e-3 &  3.05e-5 &  8.42e-3 &  7.69e-5 &  3.06e-2 &  9.42e-3 &  7.81e-3 &  2.25e-3 \\
           &      & BIM &    7.81e-3 &  3.05e-5 &  2.95e-2 &   7.7e-3 &  4.12e-2 &  3.05e-5 &  4.09e-2 &  1.01e-3 &  1.04e-2 &  3.63e-3 \\
           &      & DeepFool &    7.82e-3 &  3.16e-5 &  5.59e-2 &  3.50e-4 &  7.15e-2 &  9.23e-3 &  7.93e-2 &  5.22e-3 &  7.15e-2 &  4.73e-5 \\
           &      & CW &    7.81e-3 &  3.05e-5 &  1.01e-1 &  5.92e-3 &  9.72e-2 &  5.53e-4 &  9.64e-2 &  9.35e-3 &  7.81e-3 &   4.6e-3 \\
    \cline{2-13}
           & \multirow{4}{*}{CIFAR-100} & FGSM &    7.81e-3 &  9.77e-5 &  1.93e-2 &  3.23e-5 &  2.54e-2 &  5.58e-4 &   4.6e-2 &  1.01e-3 &  4.29e-2 &  1.28e-3 \\
           &      & BIM &    7.81e-3 &  3.12e-5 &   1.2e-1 &  1.07e-3 &  1.29e-1 &  1.10e-4 &  6.90e-2 &  2.36e-3 &  1.18e-1 &  6.21e-4 \\
           &      & DeepFool &    7.81e-3 &  3.05e-5 &  1.10e-1 &  3.28e-3 &  3.71e-2 &  9.03e-3 &  7.24e-2 &  3.05e-5 &  1.58e-1 &  5.43e-5 \\
           &      & CW &    7.81e-3 &  3.05e-5 &  1.54e-2 &  4.32e-2 &  1.44e-1 &  3.05e-5 &  1.08e-1 &  4.46e-5 &  4.77e-2 &  2.35e-3 \\
    \cline{2-13}
           & \multirow{4}{*}{SVHN} & FGSM &    7.81e-3 &  3.05e-5 &  2.81e-2 &   1.2e-4 &  4.36e-2 &  3.05e-5 &  5.53e-2 &  1.02e-3 &  2.03e-2 &  1.33e-4 \\
           &      & BIM &    7.81e-3 &  3.05e-5 &  1.19e-1 &  3.76e-2 &   1.4e-1 &  1.82e-2 &  6.88e-2 &  1.18e-3 &  5.65e-2 &  1.33e-4 \\
           &      & DeepFool &    7.81e-3 &  3.05e-5 &  7.81e-3 &  3.05e-5 &  1.78e-2 &  1.12e-2 &  1.22e-1 &  1.17e-3 &   2.4e-2 &  4.12e-5 \\
           &      & CW &    7.81e-3 &  3.05e-5 &  7.81e-3 &  4.63e-2 &  1.58e-1 &  1.18e-2 &  1.09e-2 &  9.82e-3 &  4.63e-2 &  1.62e-4 \\
    \bottomrule
    \end{tabular}
    \end{center}
    \caption{\textbf{Best hyperparameters for the OCSVM detector.} Optimal OCSVM hyperparameters ($\nu$,  $\gamma$) for all the combinations of layer, model (DenseNet, ResNet), dataset (CIFAR-10, CIFAR-100, SVHN) and attack (FGSM, BIM, DeepFool, CW).}
    \label{supp_tab:ocsvm_best_hyp}
\end{sidewaystable*}

\begin{table*}[hb]
    \centering
    \begin{minipage}{.45\linewidth}
    \begin{center}
    \footnotesize
    \begin{tabular}{ccccccccccccc}
    \toprule
             &      & {} & \multicolumn{2}{c}{Best Value} \\
             &      & Parameter & $\epsilon$ & $k$ \\
    Model & Dataset & Attack &            &     \\
    \midrule
    \multirow{12}{*}{DenseNet} & \multirow{4}{*}{CIFAR-10} & FGSM &      0.001 &  60 \\
             &      & BIM &        0.0 &  90 \\
             &      & DeepFool &        0.0 &  20 \\
             &      & CW &        0.0 &  20 \\
    \cline{2-5}
             & \multirow{4}{*}{CIFAR-100} & FGSM &     0.0014 &  90 \\
             &      & BIM &     0.0014 &  90 \\
             &      & DeepFool &        0.0 &  80 \\
             &      & CW &        0.0 &  70 \\
    \cline{2-5}
             & \multirow{4}{*}{SVHN} & FGSM &     0.0005 &  90 \\
             &      & BIM &      0.001 &  80 \\
             &      & DeepFool &     0.0005 &  20 \\
             &      & CW &        0.0 &  20 \\
    \midrule
    \multirow{12}{*}{ResNet} & \multirow{4}{*}{CIFAR-10} & FGSM &      0.001 &  80 \\
           &      & BIM &     0.0005 &  90 \\
           &      & DeepFool &      0.001 &  20 \\
           &      & CW &        0.0 &  50 \\
    \cline{2-5}
           & \multirow{4}{*}{CIFAR-100} & FGSM &     0.0005 &  90 \\
           &      & BIM &      0.001 &  90 \\
           &      & DeepFool &      0.001 &  80 \\
           &      & CW &      0.001 &  90 \\
    \cline{2-5}
           & \multirow{4}{*}{SVHN} & FGSM &     0.0005 &  80 \\
           &      & BIM &     0.0005 &  30 \\
           &      & DeepFool &      0.001 &  20 \\
           &      & CW &        0.0 &  20 \\
    \bottomrule
    \end{tabular}
    \end{center}
    \end{minipage}
    \bigskip
    \caption{\textbf{Best hyperparameters for Mahalanobis and LID detectors.} Optimal Mahalanobis and LID hyperparameters, i.e. perturbation magnitude $\epsilon$ and number of neighbors $k$, respectively, for all the combinations of method, model (DenseNet, ResNet), dataset (CIFAR-10, CIFAR-100, SVHN) and attack (FGSM, BIM, DeepFool, CW).}
    \label{supp_tab:maha_lid_best_hyp}
\end{table*}

\begin{sidewaystable*}[tb]
\begin{center}
\small
\begin{tabular}{ccccccccccc}
\toprule
       &      & Attack & \multicolumn{2}{c}{FGSM} & \multicolumn{2}{c}{BIM} & \multicolumn{2}{c}{DeepFool} & \multicolumn{2}{c}{CW} \\
       &      & {} &            AUPR &           AUROC &            AUPR &           AUROC &            AUPR &           AUROC &            AUPR &           AUROC \\
Model & Dataset & Detector &                 &                 &                 &                 &                 &                 &                 &                 \\
\midrule
\multirow{21}{*}{DenseNet} & \multirow{7}{*}{CIFAR-10} & LID &           96.37 &           98.30 &           99.52 &           99.73 &           75.21 &           85.22 &           69.75 &           80.88 \\
       &      & Maha &           99.80 &  \textbf{99.96} &           99.46 &           99.75 &           74.46 &           82.73 &           78.43 &           87.42 \\
       &      & OCSVM &           99.58 &           99.88 &           99.24 &           99.69 &           77.01 &           84.74 &           82.98 &           90.24 \\
       &      & Maha+LID &           99.69 &           99.89 &           99.83 &           99.93 &           81.11 &           88.49 &           81.22 &           89.11 \\
       &      & OCSVM+LID &           99.79 &           99.95 &  \textbf{99.90} &  \textbf{99.96} &           83.17 &           89.00 &  \textbf{85.30} &  \textbf{91.50} \\
       &      & OCSVM+Maha &  \textbf{99.85} &           99.94 &           99.47 &           99.78 &           79.78 &           86.52 &           83.17 &           90.49 \\
       &      & EAD &           99.70 &           99.89 &           \textbf{99.90} &           \textbf{99.96} &  \textbf{83.70} &  \textbf{89.36} &           \textbf{85.30} &           \textbf{91.50} \\
\cline{2-11}
       & \multirow{7}{*}{CIFAR-100} & LID &           98.39 &           99.29 &           96.31 &           98.11 &           55.88 &           70.12 &           59.67 &           72.80 \\
       &      & Maha &           99.49 &           99.87 &           97.64 &           99.10 &           67.79 &           78.49 &           74.99 &           86.86 \\
       &      & OCSVM &           99.49 &           99.84 &           98.23 &           99.30 &           69.17 &           79.27 &           78.71 &           88.95 \\
       &      & Maha+LID &           99.69 &           99.90 &           97.85 &           99.11 &           70.86 &           81.02 &           79.06 &           89.82 \\
       &      & OCSVM+LID &  \textbf{99.80} &  \textbf{99.93} &  \textbf{98.85} &  \textbf{99.57} &  \textbf{73.24} &           83.06 &           82.32 &           91.78 \\
       &      & OCSVM+Maha &           99.62 &           99.89 &           98.15 &           99.37 &           69.71 &           80.07 &           81.37 &           90.20 \\
       &      & EAD &           99.78 &           \textbf{99.93} &           98.37 &           99.47 &           73.05 &  \textbf{83.07} &  \textbf{83.40} &  \textbf{92.15} \\
\cline{2-11}
       & \multirow{7}{*}{SVHN} & LID &           98.59 &           99.07 &           92.13 &           94.79 &           85.80 &           91.83 &           90.47 &           94.61 \\
       &      & Maha &           99.45 &           99.85 &           97.93 &           99.26 &           90.00 &           94.93 &           90.95 &           97.16 \\
       &      & OCSVM &           99.51 &           99.86 &           97.38 &           99.17 &           91.40 &           95.00 &           96.54 &           98.50 \\
       &      & Maha+LID &           99.80 &  \textbf{99.93} &           98.64 &           99.50 &           92.15 &           95.65 &           95.29 &           98.32 \\
       &      & OCSVM+LID &           99.81 &           99.93 &           98.45 &           99.44 &           92.63 &           95.58 &  \textbf{98.21} &  \textbf{99.19} \\
       &      & OCSVM+Maha &           99.54 &           99.87 &           98.50 &           99.42 &           92.38 &           95.66 &           96.97 &           98.65 \\
       &      & EAD &  \textbf{99.83} &           99.91 &  \textbf{98.93} &  \textbf{99.57} &  \textbf{93.27} &  \textbf{96.04} &           98.13 &           99.16 \\
\bottomrule
\end{tabular}
\end{center}
\caption{{\bf Comparative assessment of \alg{} and competing methods (DenseNet).} Performance comparison of the \alg{},  LID \cite{maCharacterizingAdversarialSubspaces2018}, Mahalanobis  \cite{leeSimpleUnifiedFramework2018}, OCSVM detectors, and of the pairwise integration of the three single detectors. The table returns the AUROC and AUPR for the DenseNet model and all the combinations of selected datasets (CIFAR-10, CIFAR-100 and SVHN) and attacks (FGSM, BIM, DeepFool and CW). See Methods for further details. }
\label{supp_tab:pairwise_results}
\end{sidewaystable*}

\begin{sidewaystable*}[tb]
\begin{center}
\small
\begin{tabular}{ccccccccccc}
\toprule
       &      & Attack & \multicolumn{2}{c}{FGSM} & \multicolumn{2}{c}{BIM} & \multicolumn{2}{c}{DeepFool} & \multicolumn{2}{c}{CW} \\
       &      & {} &            AUPR &           AUROC &            AUPR &           AUROC &            AUPR &           AUROC &            AUPR &           AUROC \\
Model & Dataset & Detector &                 &                 &                 &                 &                 &                 &                 &                 \\
\midrule
\multirow{21}{*}{ResNet} & \multirow{7}{*}{CIFAR-10} & LID &           99.18 &           99.67 &           94.37 &           96.50 &           79.40 &           88.58 &           73.95 &           82.29 \\
       &      & Maha &           99.87 &           99.90 &           99.06 &           99.58 &           85.64 &           91.60 &           92.28 &           95.90 \\
       &      & OCSVM &  \textbf{99.99} &  \textbf{99.99} &           98.95 &           99.44 &           83.90 &           90.83 &           92.26 &           95.68 \\
       &      & Maha+LID &           99.97 &           99.98 &           99.45 &           99.73 &           86.40 &           92.26 &           92.53 &           96.07 \\
       &      & OCSVM+LID &           \textbf{99.99} &           \textbf{99.99} &           99.54 &           99.75 &           85.36 &           91.71 &           92.81 &           96.08 \\
       &      & OCSVM+Maha &           \textbf{99.99} &           \textbf{99.99} &           99.34 &           99.67 &  \textbf{87.95} &           92.45 &           93.24 &           96.36 \\
       &      & EAD &           \textbf{99.99} &           \textbf{99.99} &  \textbf{99.58} &  \textbf{99.78} &           87.77 &  \textbf{92.89} &  \textbf{93.35} &  \textbf{96.46} \\
\cline{2-11}
       & \multirow{7}{*}{CIFAR-100} & LID &           97.53 &           98.78 &           94.52 &           96.76 &           56.10 &           69.87 &           65.53 &           78.51 \\
       &      & Maha &           99.48 &           99.72 &           93.51 &           96.92 &           73.32 &           85.23 &           83.00 &           91.68 \\
       &      & OCSVM &  \textbf{99.63} &  \textbf{99.86} &           91.70 &           95.79 &           71.69 &           84.17 &           83.17 &           91.24 \\
       &      & Maha+LID &           99.58 &           99.79 &           97.63 &           98.94 &           74.66 &           85.65 &           83.56 &           92.50 \\
       &      & OCSVM+LID &           99.21 &           99.65 &  \textbf{98.24} &           99.13 &           73.64 &           85.18 &           84.43 &           92.74 \\
       &      & OCSVM+Maha &           99.63 &           99.78 &           94.48 &           97.63 &           76.24 &           85.74 &           87.16 &           93.01 \\
       &      & EAD &           \textbf{99.63} &           99.78 &           98.22 &  \textbf{99.26} &  \textbf{76.58} &  \textbf{86.34} &  \textbf{88.26} &  \textbf{94.08} \\
\cline{2-11}
       & \multirow{7}{*}{SVHN} & LID &           94.52 &           97.84 &           83.46 &           90.78 &           86.60 &           92.31 &           79.46 &           88.16 \\
       &      & Maha &           97.90 &           99.60 &           92.22 &           97.16 &           93.04 &           95.74 &           84.95 &           92.13 \\
       &      & OCSVM &           98.06 &           99.64 &           95.91 &           98.12 &           92.15 &           95.58 &           89.19 &           93.29 \\
       &      & Maha+LID &           98.05 &           99.66 &           93.71 &           97.74 &           93.52 &           96.09 &           87.43 &           93.60 \\
       &      & OCSVM+LID &           98.12 &  \textbf{99.69} &  \textbf{96.84} &           98.58 &           93.03 &           95.96 &  \textbf{90.88} &           94.59 \\
       &      & OCSVM+Maha &  \textbf{98.39} &           99.68 &           95.83 &           98.14 &           93.48 &           96.00 &           89.19 &           93.41 \\
       &      & EAD &           98.33 &           \textbf{99.69} &           96.80 &  \textbf{98.59} &  \textbf{93.70} &  \textbf{96.18} &           90.66 &  \textbf{94.62} \\
\bottomrule
\end{tabular}
\end{center}
\caption{{\bf Comparative assessment of \alg{} and competing methods (ResNet).} Performance comparison of the \alg{},  LID \cite{maCharacterizingAdversarialSubspaces2018}, Mahalanobis  \cite{leeSimpleUnifiedFramework2018}, OCSVM detectors, and of the pairwise integration of the three single detectors. The table returns the AUROC and AUPR for the ResNet model and all the combinations of selected datasets (CIFAR-10, CIFAR-100 and SVHN) and attacks (FGSM, BIM, DeepFool and CW). See Methods for further details. }
\label{supp_tab:pairwise_results}
\end{sidewaystable*}

\printbibliography